\documentclass[10pt,journal]{IEEEtran}
\usepackage{enumitem,calc}
\usepackage{graphicx}
\usepackage{subfigure}
\usepackage{amsmath}
\usepackage{array}
\usepackage{subfig}
\usepackage{tabu}
\usepackage{amssymb}
\usepackage{multirow}
\usepackage{booktabs}
\usepackage{graphicx}

\usepackage{cite}
\usepackage{xcolor}
\usepackage{amsfonts}
\usepackage{makecell}
\usepackage[noend]{algpseudocode}
\usepackage{soul}
\soulregister\cite7
\soulregister\ref7
\soulregister\eqref7
\ifCLASSINFOpdf
\else
\fi
\begin{document}
%
\title{RDNet: Region Proportion-Aware  Dynamic Adaptive Salient Object Detection Network in Optical Remote Sensing  Images}

\author{Bin~Wan,
	Runmin~Cong,
	Xiaofei~Zhou,
	Hao~Fang,
	Yaoqi~Sun,
	and Sam~Kwong, \emph{Fellow}, \emph{IEEE}
		\thanks{This work was supported  in part by  the opening project of State Key Laboratory of Autonomous Intelligent Unmanned Systems under Grant ZZKF2025-2-8, part by the Taishan Scholar Project of Shandong Province under Grant tsqn202306079, in part by the National Natural Science Foundation of China Grant 62471278 and 62271180, and in part by the Research Grants Council of the Hong Kong Special Administrative Region, China under Grant STG5/E-103/24-R.\emph{(Corresponding author: Runmin Cong)}.}
	\thanks{ Bin Wan, Runmin Cong and Hao Fang are with the School of Control Science and Engineering, Shandong University, Jinan 250061, China, and also with the Key Laboratory of Machine Intelligence and System Control, Ministry of Education, Jinan 250061, China. (E-mail:  wanbinxueshu@icloud.com; rmcong@sdu.edu.cn; fanghaook@mail.sdu.edu.cn).}
	
	\thanks{ Xiaofei Zhou is with School of Automation, Hangzhou Dianzi University, Hangzhou 310018, China (E-mail:  zxforchid@outlook.com).}
	\thanks{Yaoqi Sun is with School of Mathematics and Computer Science, Lishui University and Lishui Institute of Hangzhou Dianzi University, Hangzhou 310018, China (E-mail: sunyq2233@163.com).}
	
	\thanks{ Sam Kwong is with the School of Data Science, Lingnan University, Tuen
		Mun, Hong Kong (E-mail: samkwong@ln.edu.hk).}
	
		
	
}

\maketitle

\begin{abstract}
	Salient object detection (SOD) in remote sensing images faces significant challenges due to large variations in object sizes, the computational cost of self-attention mechanisms, and the limitations of CNN-based extractors in capturing global context and long-range dependencies. Existing methods that rely on fixed convolution kernels often struggle to adapt to diverse object scales, leading to detail loss or irrelevant feature aggregation. To address these issues, this work aims to enhance robustness to scale variations and achieve precise object localization. We propose the Region Proportion-Aware Dynamic Adaptive Salient Object Detection Network (RDNet), which replaces the CNN backbone with the SwinTransformer for global context modeling and introduces three key modules: (1) the Dynamic Adaptive Detail-aware (DAD) module, which applies varied convolution kernels guided by object region proportions; (2) the Frequency-matching Context Enhancement (FCE) module, which enriches contextual information through wavelet interactions and attention; and (3) the Region Proportion-aware Localization (RPL) module, which employs cross-attention to highlight semantic details and integrates a Proportion Guidance (PG) block to assist the DAD module. By combining these modules, RDNet achieves robustness against scale variations and accurate localization, delivering superior detection performance compared with state-of-the-art methods.

\end{abstract}



\IEEEpeerreviewmaketitle

\section{Introduction}
\IEEEPARstart{W}{hen} the human eyes observe a natural scene, it typically directs its attention first to certain prominent areas-such as regions with vivid colors, geometric anomalies, kinetic motion patterns, or strong contrast. This phenomenon, commonly referred to as ``human visual attention", plays a pivotal role in the process of visual perception. To emulate this mechanism, numerous researchers in the field of computer vision have devoted themselves to designing salient object detection (SOD) methods, which aim to locate and delineate the most visually prominent objects or regions within a scene. By focusing on these salient areas, such methods can effectively suppress irrelevant background clutter, thereby reducing computational overhead and improving detection accuracy. Consequently, salient object detection has found wide-ranging applications across multiple domains\cite{cong2025breaking,qin2025sight,cong2025divide,cong2025trnet,cong2024query}, including but not limited to defect detection \cite{huang2022small,wan2023lfrnet}, camouflaged object detection \cite{zhuge2022cubenet,chen2022camouflaged}, semantic object segmentation\cite{chen2025replay,cong2025generalized,chen2025empowering,xiong2025mm,cong2025uis,chen2025replay,lian2024diving} and light field object detection \cite{zhang2020lfnet,piao2021panet,jing2021occlusion}.

In recent years, the advent of convolutional neural networks (CNNs) \cite{cong2025reference,fang2025decoupled,cong2022bcs} has expanded the application of salient object detection to remote sensing images \cite{ren2022ship,zeng2023adaptive,yao2024iterative}, substantially enhancing detection accuracy through CNNs' robust and efficient feature extraction capabilities. For example, in \cite{li2023lightweight}, Li \emph{et al}. adopted CNN-based MobileNet-V2 to extract hierarchical  features, and designed the  dynamic semantic matching module and edge self-alignment module to optimize high-level semantic features and low-level detail features. In \cite{zeng2023adaptive}, Zeng \emph{et al}. used the VGG as the backbone network and designed the multiscale feature extraction module with multiple dilated convolutional layers to capture and utilize multiscale information. In \cite{li2023salient}, Li \emph{et al}. employed a direction-aware shuffle weighted spatial attention module and its simplified version to enhance local interactions, and a knowledge transfer module  to further enhance cross-level contextual interactions. 

\begin{figure}[t]
	\centering
	\includegraphics[width=0.4\textwidth]{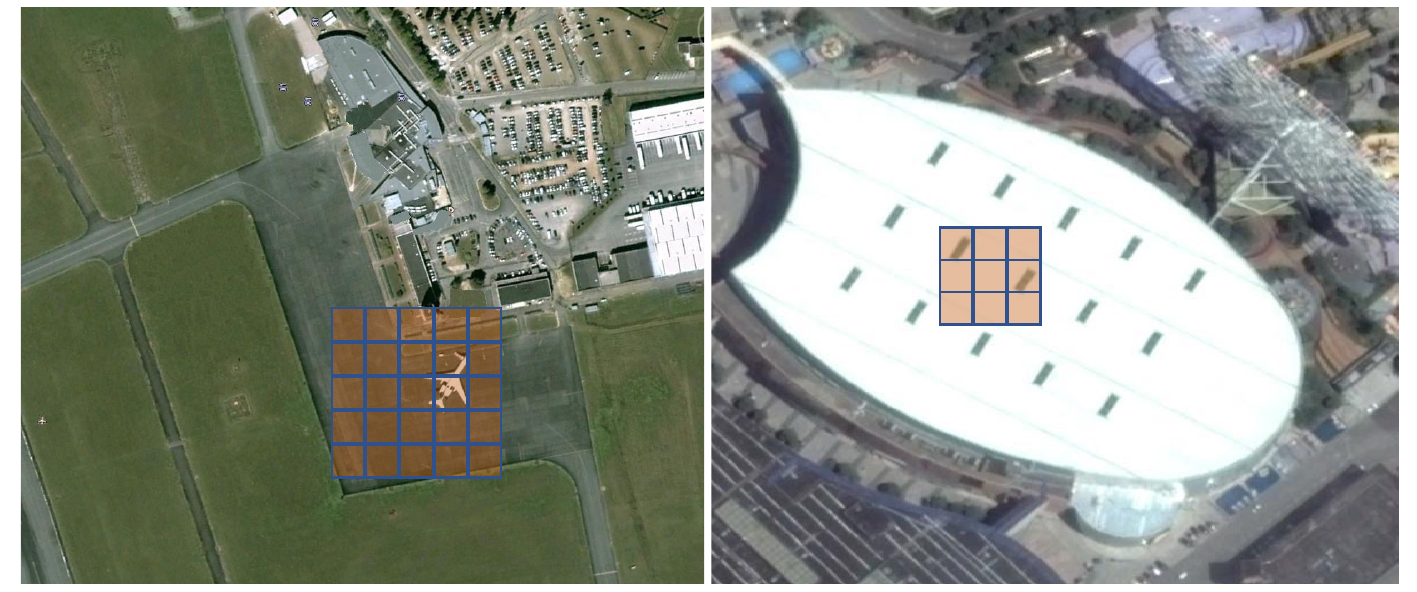}
	\caption{\small{Influence of convolution kernels of different sizes.}}
	\label{fig_ker}
\end{figure}

However, due to the inherent characteristics of remote sensing scene images, where object sizes vary significantly, most methods that use combinations of different-sized convolution kernels to extract object details cause the network to either overlook the overall region or to over-focus on irrelevant areas. Consequently, as illustrated in Fig. \ref{fig_ker}, when the object is too small, large convolution kernels integrate excessive background information, and when the object region is too large, small convolution kernels fail to capture the entire region of the object. Besides, to explore contextual interactions between inter-level features, most methods \cite{gongyangli2022lightweight,li2023salient} adopt self-attention mechanism \cite{vaswani2017attention} performs a matrix multiplication of adjacent features at their full resolution to achieve feature interaction, which not only significantly increases computational overhead but also directly merges high- and low-frequency information, resulting in diluted object information. In addition, CNN-based feature extractors rely on local convolution kernels, which can limit their capacity to capture global context and handle long-range dependencies.

To overcome the limitations mentioned above, in this paper, we propose a Region Proportion-aware  Dynamic Adaptive  Salient Object Detection Network which provides a new solution for the ORSI-SOD, our main idea is that replace the CNN with a Transformer as the feature extractor to capture global context information. On this basis, we design the dynamic adaptive  detail-aware (DAD) module, frequency-matching context enhancement (FCE) module and region proportion-aware localization (RPL) module to extract the detail, contextual  and location information from the multi-layer features. To be specific, the SwinTransformer is selected as the backbone network. First,  to address the challenge of random object locations in remote sensing images, we integrate the RPL module  into the high-level features, through continuous cross-attention operations, the network is guided to concentrate on location information. Besides, considering that the impact of large variations in object scale within remote sensing images, we introduce a proportion guidance (PG) block in the RPL module to compute the object region proportion through global average pooling and fully connected layers, and then provides guidance for subsequent DAD module. After that, unlike the earlier approach of using the same-sized convolution kernels for all features, the DAD module employs differently sized kernels based on varying regional proportions, enabling the extraction of object information across diverse receptive fields. In addition, considering that directly applying self-attention mechanisms to full-resolution features significantly increases computational overhead and introduces interference between low-frequency and high-frequency information, to further explore the  contextual information, we design a FCE module with both a wavelet interaction stage and a feature enhancement stage. Instead of directly relying on self-attention for feature interaction between adjacent layers, in the wavelet interaction stage, the corresponding frequency components interact to extract richer contextual information. During the feature enhancement stage, channel and spatial attention mechanisms are applied to achieve refined information filtering. Finally, by fusing the output features of the three modules in a bottom-up manner, high-quality detection results are obtained. 


The main contributions of this paper are summarized as follows,
\begin{enumerate}[leftmargin=*]
\item We propose a novel Region Proportion-aware  Dynamic Adaptive  Salient Object Detction Network in Optical Remote Sensing  Images which includes dynamic adaptive  detail-aware (DAD) module, frequency-matching context enhancement (FCE)  module and region proportion-aware localization (RPL) module. Experimental results on  three public remote sensing  image dataset prove that our RDNet outperforms the other state-of-the-art saliency detection methods.
\item  We propose the dynamic adaptive  detail-aware (DAD) module which dynamically selects the combinations of convolution kernels of different sizes according to different regional proportions to extract the object detail information.
\item We design the frequency-matching context enhancement (FCE) module which consists of wavelet interaction stage and  feature enhancement stage to extract and optimize the context features.
\item We design the region proportion-aware localization (RPL) module which leverages cross-attention operations to mine the location information embedded in the high-level features and introduces a proportion guidance (PG) block to provide cues for DAD module.
\end{enumerate}

The rest of this paper is organized as follows. The related works are reviewed in Section II. The proposed RDNet is described in Section III. Experimental results and the related analyses are presented in Section IV. Finally, we conclude our method in Section V.
\begin{figure*}[t]
\centering
\includegraphics[width=0.9\textwidth]{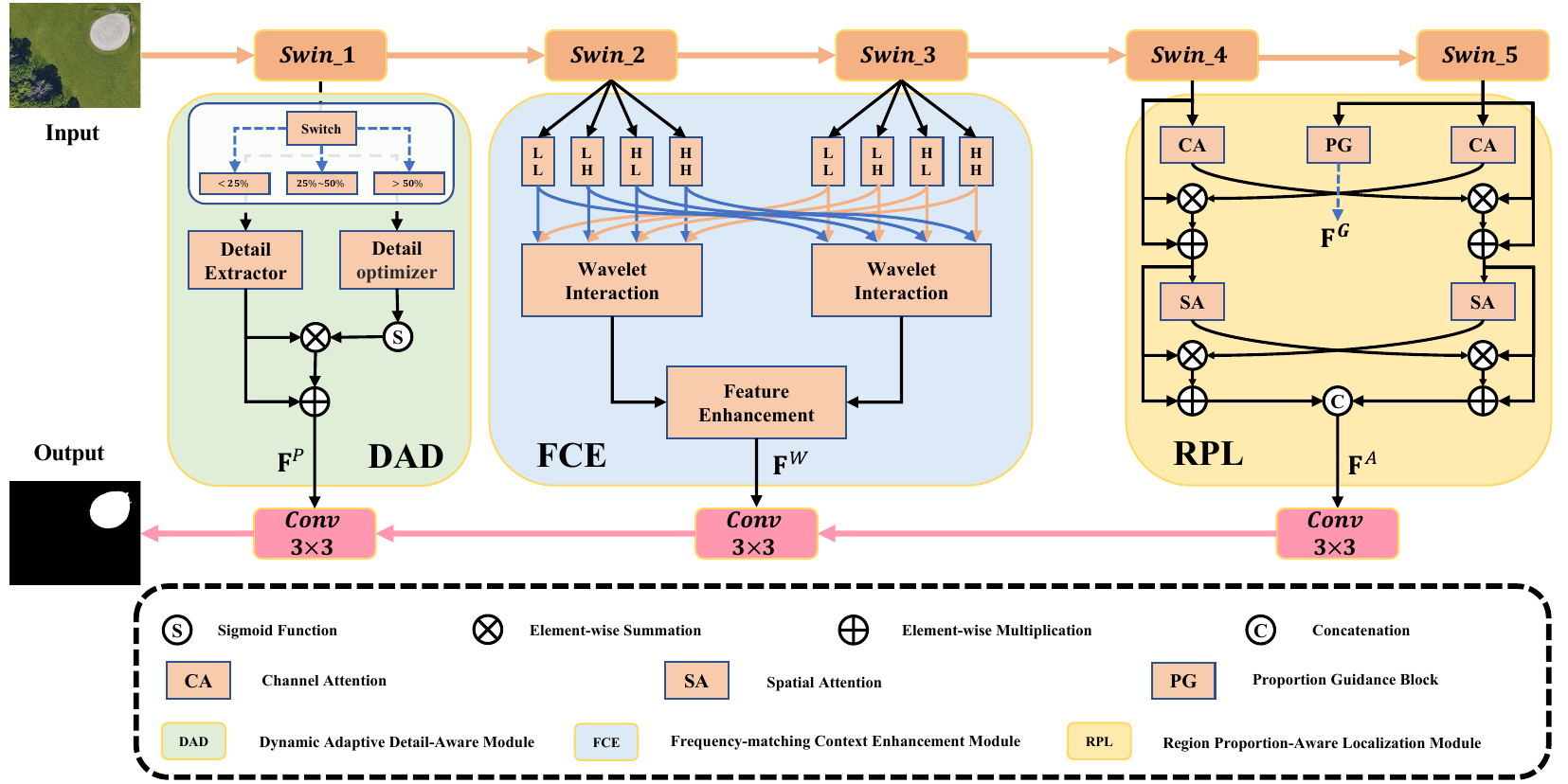}
\caption{\small{The overall architecture of proposed RDNet.}}
\label{fig_ov}
\end{figure*}


\section{Related Works}
In this section, we briefly review the salient object detection, including the  salient object detection for natural scene Images and ORSI-SOD.

\subsection{Salient Object Detection for Natural Scene Images}
In the last two decades, many efforts have been spent in the salient object detection for natural scene Images. Initially, traditional methods  based on various hand-crafted features \cite{li2016single} (\emph{e.g}., image contrast, pixel brightness, color) were proposed to deal with various natural scenes. For example, in \cite{ma2003contrast}, Ma \emph{et al}. leveraged the contrast analysis and image attention analysis to construct  a feasible and fast approach. In \cite{han2006unsupervised}, Han \emph{et al}. proposed an unsupervised model for extracting attention objects from color images by combining visual attention mechanisms with object-growing techniques. Recently, deep learning techniques have recently made remarkable strides in SOD tasks. In \cite{li2020complementarity}, Li \emph{et al}. introduced a complementarity-aware attention network that jointly detects foreground and background regions using two branches: a positive attention module  and a negative attention module. In \cite{fang2023udnet}, Fang \emph{et al}. proposed an uncertainty-aware SOD model that addresses ambiguous predictions around object contours using multiple supervision signals and a feature interaction module. In \cite{lu2024low}, Lu \emph{et al}. designed a two-branch framework for salient object detection in low-light images, combining adaptive foreground enhancement and refined detection.

Although salient object detection (SOD) methods designed for natural scene images (NSI) have achieved substantial progress, directly applying these approaches to remote sensing imagery (RSI) remains non-trivial. This difficulty arises from the distinct characteristics of RSI, including more complex backgrounds, diverse object categories, varied imaging angles. Such differences result in a significant domain gap that hampers the effectiveness of NSI-based SOD methods when transferred to the RSI domain. Nonetheless, these NSI-SOD approaches provide valuable methodological insights and serve as an important foundation for advancing salient object detection techniques tailored specifically to the unique challenges of RSI.
\subsection{Salient Object Detection for ORSI}
Following the introduction of saliency models for natural scene images, this section presents a brief review of CNN-based methods for salient object detection in optical remote sensing images (ORSI-SOD). 

Li \cite{li2019nested} \emph{et al}. constructed the first publicly available optical RSI dataset and proposed a deep network for salient object detection in optical remote sensing images which tackles challenges like scale variation and cluttered backgrounds. In subsequent work, Zhang \cite{zhang2020dense} and Tu \cite{tu2021orsi} introduced two datasets EORSSD and ORSI-4199 aimed at improving detection performance in remote sensing image. Since then, numerous researchers have devoted themselves to ORSI-SOD. For example, in \cite{wang2022hybrid}, Wang \emph{et al}. proposed a hybrid feature-aligned network for salient object detection in optical remote sensing images, addressing challenges like cluttered backgrounds, scale variation, and irregular object edges. In \cite{liu2023distilling}, Liu \emph{et al}. designed a super-resolution-assisted learning framework which reduces input resolution and integrates a transposed decoder , auxiliary SR decoder, and task-fusion guidance module for efficient and accurate salient object detection in remote sensing images. In \cite{yao2024iterative}, Yao \emph{et al}. combined  iterative saliency aggregation and assignment with adaptive feature fusion to balance accuracy and efficiency for remote sensing saliency detection. In \cite{yan2024asnet}, Yan  \emph{et al}. proposed an adaptive semantic network  or salient object detection in optical remote sensing images by combining Transformer and CNN in a dual-branch encoder, and introducing ASMM, AFEM, and MFIM modules to effectively capture global-local features and enhances saliency precision. In \cite{zhao2024recurrent}, Zhao \emph{et al}. addressed complex object structures in ORSI-SOD by jointly modeling regions and boundaries through graph reasoning.

Despite the encouraging progress made by the aforementioned methods, several critical challenges in optical remote sensing imagery (RSI) remain insufficiently addressed. In particular, the larger variations in object scales, the irregular topological structures of salient objects and the interference caused by complex and cluttered backgrounds continue to hinder accurate detection. To tackle these issues, we introduce a novel framework, RDNet, which is specifically designed to exploit heterogeneous feature representations and adopts distinct feature optimization strategies tailored to objects of varying scales. The architectural details and key components of our proposed model are thoroughly described in Section III.

\section{Proposed Framework}
\subsection{Overview of Proposed RDNet}
In this paper, we propose a Region Proportion-aware  Dynamic Adaptive Salient Object Detction Network (RDNet) in  Optical Remote Sensing  Images which consists of dynamic adaptive  detail-aware (DAD) module, frequency-matching context enhancement (FCE)  module and region proportion-aware localization (RPL) module. Concretely, we use the SwinTransformer as the backbone network, whose input is set to $4\times3\times384\times384$, to extract the multi-level feature  $\{\mathbf{F}_i^{R}\}_{i=1}^5$. Firstly, considering that high-level features $\mathbf{F}_4^{R}$ and $\mathbf{F}_5^{R}$ contain more location information, the RPL module adopts  continuous cross-attention operations to complete feature optimization, yielding location feature $\mathbf{F}^{A}$. Besides, in order to calculate the proportion of the object area in the entire image, we introduce a PG block in the RPL module that utilizes global average pooling and fully connected layers to obtain the features $\mathbf{F}^{G}$ $\in \mathbb{R}^{4\times1\times1\times1} $. Then, different from previous methods which leverage the same-sized convolution kernels for all features, the DAD module which is guided by $\mathbf{F}^{G}$ to dynamically select combinations of different-sized convolution kernels, supplemented by an attention mechanism, to obtain detail feature $\mathbf{F}^{P}$ based on varying area proportions. After that, features $\mathbf{F}_2^{R}$ and $\mathbf{F}_3^{R}$ are fed into the FCE module, where the discrete wavelet transform is utilized to generate four frequency components (\emph{i.e.}, $\mathbf{F}_{2/3}^{ll}$,$\mathbf{F}_{2/3}^{lh}$,$\mathbf{F}_{2/3}^{hl}$,$\mathbf{F}_{2/3}^{hh}$)  and the wavelet interaction block is deployed to achieve feature interaction between adjacent layers, the attention operations are applied to interaction features to gain context feature $\mathbf{F}^{W}$.  Finally, features $\mathbf{F}^{A}$, $\mathbf{F}^{W}$ and $\mathbf{F}^{P}$ are integrated in a bottom-top manner to generate the final saliency map $\mathbf{S}$.

\subsection{Region Proportion-aware Localization Module}
To fully explore and utilize the semantic information in high-level features, taking feature $\mathbf{F}_4^{R} $ and feature $\mathbf{F}_5^{R}$  as inputs, we propose an region proportion-aware localization (RPL) module, where the location information is optimized through consecutive cross-attention operations. As shown in Fig. \ref{fig_ov}, first, the channel attention operation is applied to $\mathbf{F}_4^{R} $ and $\mathbf{F}_5^{R}$ for generating two feature vectors $\mathbf{V}_4 $ and $\mathbf{V}_5$ which are respectively subjected to multiplication and addition operations with the input features of the other, thereby achieving optimization in the channel dimension and yielding $\mathbf{F}_4^{ca} $ and $\mathbf{F}_5^{ca}$,
\begin{equation}
	\begin{cases}
		\mathbf{V}_4 = f_{CA}(\mathbf{F}_4^{R})\\
		\mathbf{V}_5 = f_{CA}(\mathbf{F}_5^{R})\\
		\mathbf{F}_4^{ca} = \mathbf{F}_4^{R} \otimes \mathbf{V}_4 \oplus\mathbf{F}_4^{R}\\
		\mathbf{F}_5^{ca} = \mathbf{F}_5^{R} \otimes \mathbf{V}_5 \oplus\mathbf{F}_5^{R}\\
	\end{cases},
\end{equation}
where $f_{CA}$ denotes the channel attention which consists of a global average pooling layer, two 1$\times$1 convolutional layers and a sigmoid activation layer.
After that, to optimize once again in the spatial dimension, the same operations described above are applied to features $\mathbf{F}_4^{ca} $ and $\mathbf{F}_5^{ca}$, where the channel attention is replaced by spatial attention, obtaining $	\mathbf{F}_4^{sa} $ and $	\mathbf{F}_5^{sa} $,

\begin{equation}
	\begin{cases}
		\mathbf{W}_4 = f_{SA}(\mathbf{F}_4^{ca})\\
		\mathbf{W}_5 = f_{SA}(\mathbf{F}_5^{ca})\\
		\mathbf{F}_4^{sa} = \mathbf{F}_4^{ca} \otimes \mathbf{W}_4 \oplus\mathbf{F}_4^{ca}\\
		\mathbf{F}_5^{sa} = \mathbf{F}_5^{ca} \otimes \mathbf{W}_5 \oplus\mathbf{F}_5^{ca}\\
	\end{cases},
\end{equation}
where $f_{SA}$ denotes the spatial attention which includes a global max pooling along the channel dimension and a sigmoid activation layer.

Finally, we deploy the concatenation operation and $3\times3$ convolutional layer to fuse $	\mathbf{F}_4^{sa} $ and $	\mathbf{F}_5^{sa} $ to gain output feature $	\mathbf{F}^{A} $,
\begin{equation}
	\mathbf{F}^{A} = f_{3\times3}(Cat(\mathbf{F}_4^{sa} ,\mathbf{F}_5^{sa} )).
\end{equation}


In salient object detection tasks, low-level features often contain abundant target detail information, such as shape and contour. To effectively extract this information, many methods utilize combinations of convolution kernels with varying sizes, such as applying convolutions with different dilation rates to capture object details across diverse receptive fields. However, in the remote sensing images, the scale variation of targets is exceptionally large. Using the same convolution kernel strategy for all objects can result in the aggregation of irrelevant information and the loss of overall details, as illustrated in Fig. \ref{fig_ker}. To address this challenge, inspired by classification tasks, we introduce a proportion guidance (PG) block in the RPL module for calculating the object region proportion. As shown in Fig. \ref{fig_ov}, feature $\mathbf{F}_5^{R}$ is first fed into the PG block as the input, where it undergoes global average pooling followed by two fully connected layers to produce feature $\mathbf{F}^{G}$. Each value in $\mathbf{F}^{G}$ corresponds to the regional proportion of its respective batch, providing  guidance for subsequent processing,

\begin{equation}
	\mathbf{F}^{G} = f_{fc\times2}(f_{avg}(\mathbf{F}_5^{R})),
\end{equation}
where $f_{fc\times2}$ denotes two fully connected layers, $f_{avg}$ means global average pooling. Besides, to ensure that the network accurately captures the region proportion of the target, we compute a loss between feature $\mathbf{F}^{G}$ and the ground-truth region proportion labels.

\begin{figure}[!t]
	\centering
	\includegraphics[width=0.45\textwidth]{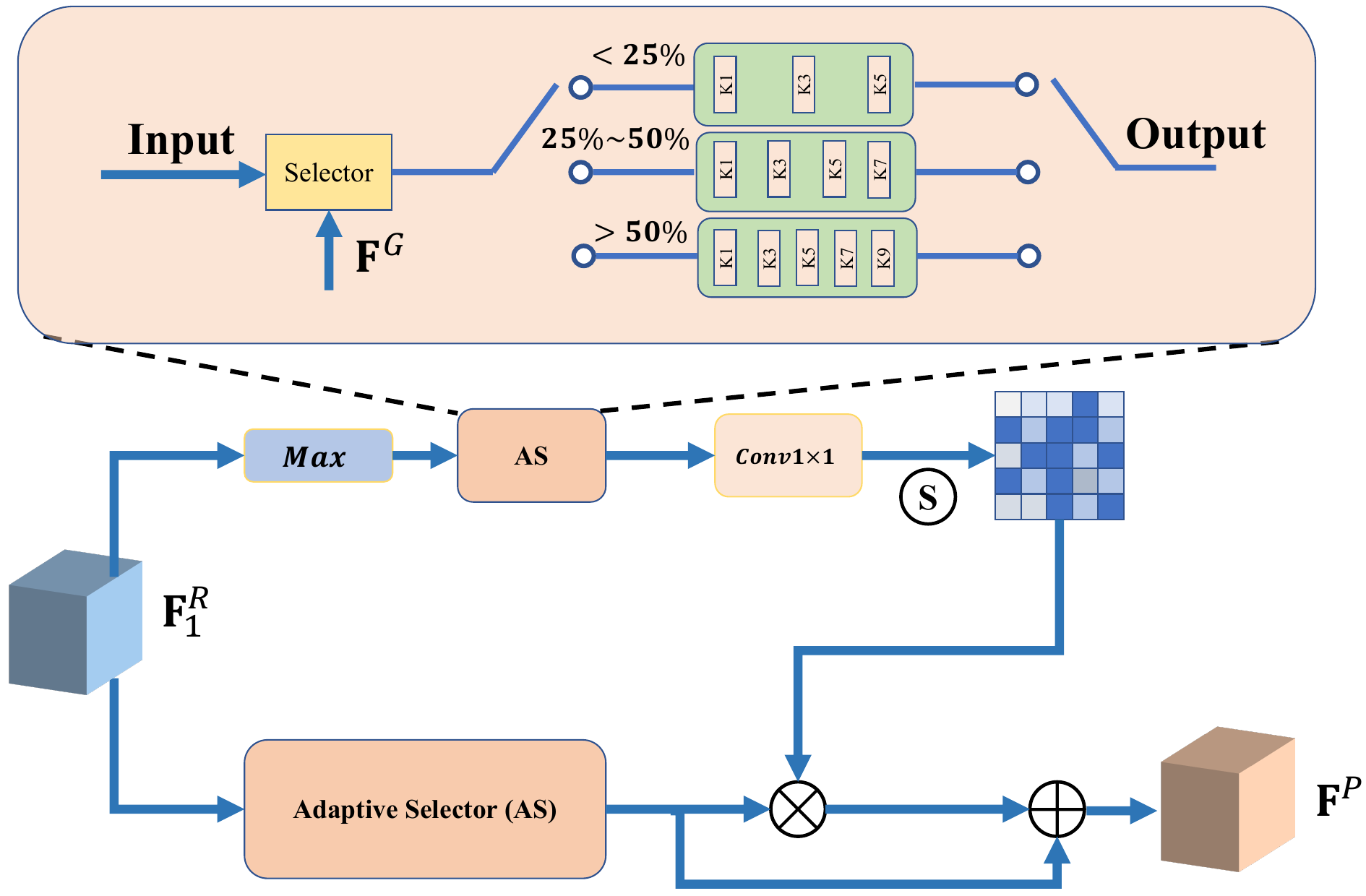}
	\caption{\small{Architecture of the dynamic adaptive  detail-aware module.}} 
	\label{fig_large}
\end{figure}

\subsection{Dynamic Adaptive  Detail-aware Module}

 After extracting location cues, to  further effectively mine detailed features, we design the  dynamic adaptive  detail-aware (DAD)  module incorporating three strategies based on the region proportion: less than 25$\%$, between 25$\%$ and 50$\%$, and greater than 50$\%$. As shown in Fig. \ref{fig_large}, the DAD module consists of two branches: the lower branch serves as a detail extractor, while the upper branch functions as a detail optimizer. Each branch includes an adaptive selector that can automatically choose different convolutional combinations based on the output of the PG block. When the object proportion exceeds 50$\%$, the DAD module applies five convolution kernels of varying sizes, processing the input features through two separate branches. Specifically, on the lower branch, considering the large proportion of the object area, larger convolution kernels (\emph{i.e.}, $7\times7$ and $9\times9$) are used to capture the overall region. Meanwhile, smaller convolution kernels (\emph{i.e.}, $3\times3$ and $5\times5$) are adopted to refine the edges of the object, addressing irregular contours. To preserve the original information of the features, a $1\times1$ convolution is applied to the input features. Finally, the outputs from the five different convolution operations are fused through addition, enabling information extraction across various receptive fields.

\begin{equation}
	\begin{cases}
		
		\mathbf{F}_1^{R_i} = f_{j\times j}(\mathbf{F}_1^{R})\\
		\mathbf{F}_1^{D} = \sum_{i=1}^{5} \mathbf{F}_1^{R_i}
		
	\end{cases} ,
\end{equation}
where $i = 1, 2, 3, 4, 5$, $f_{j\times j}$ denote the convolutonal layers with kernel size $j$ and $j = 2 \times i -1$. Besides,  considering that low-level features contain not only detailed information but also significant noise, we incorporate a spatial attention mechanism in the upper branch. Specifically, the input features first undergo global max pooling along the channel dimension to produce a feature map, which is then processed through five different convolutional layers in the adaptive selector to generate corresponding features (\emph{i.e.},	$\mathbf{W}_1^{R_1}$,$\mathbf{W}_2^{R_1}$,$\mathbf{W}_3^{R_1}$,$\mathbf{W}_4^{R_1}$,$\mathbf{W}_5^{R_1}$ ). The outputs from these five layers are summed, followed by a $1\times1$ convolution and a sigmoid activation function to compute the final feature weight $\mathbf{W}$,

\begin{equation}
	\begin{cases}
		
		\mathbf{W}_1^{R_i} = f_{j\times j}(f_{max}(\mathbf{F}_1^{R}))\\
		\mathbf{W} = \sum_{i=1}^{5} \mathbf{W}_1^{R_i}
		
	\end{cases} ,
\end{equation}
where $f_{max}$ means global max pooling along the channel dimension. Finally, to complete feature optimization,  multiplication and addition operations are applied to feature $\mathbf{F}_1^{D} $ and feature $\mathbf{W}$ to obtain the output feature  $\mathbf{F}^{P}$ of the DAD module,
\begin{equation}
    \mathbf{F}^{P} = \mathbf{F}_1^{D} \otimes \mathbf{W} \oplus \mathbf{F}_1^{D}.
\end{equation}

For objects with other regional proportions (\emph{i.e.}, $<25\%$ and $25\% \sim 50\%$), we employ  three and four convolutional operations respectively.

\subsection{Frequency-matching Context Enhancement  Module }

Considering that high-level features provide rich semantic information while low-level features offer precise positional details, we design the DAD module and the RPL module to effectively extract and optimize these features. For middle-layer features, which carry important contextual information, most existing methods rely on applying self-attention mechanisms directly to full-resolution features. However, this approach not only greatly increases computational overhead but also leads to interference between low-frequency and high-frequency information, compromising the overall feature quality.  Inspired by the wavelet transformer, we propose the frequency-matching context enhancement (FCE) module which consists of wavelet interaction stage and feature enhancement stage.
\begin{figure}[!t]
	\centering
	\includegraphics[width=0.49\textwidth]{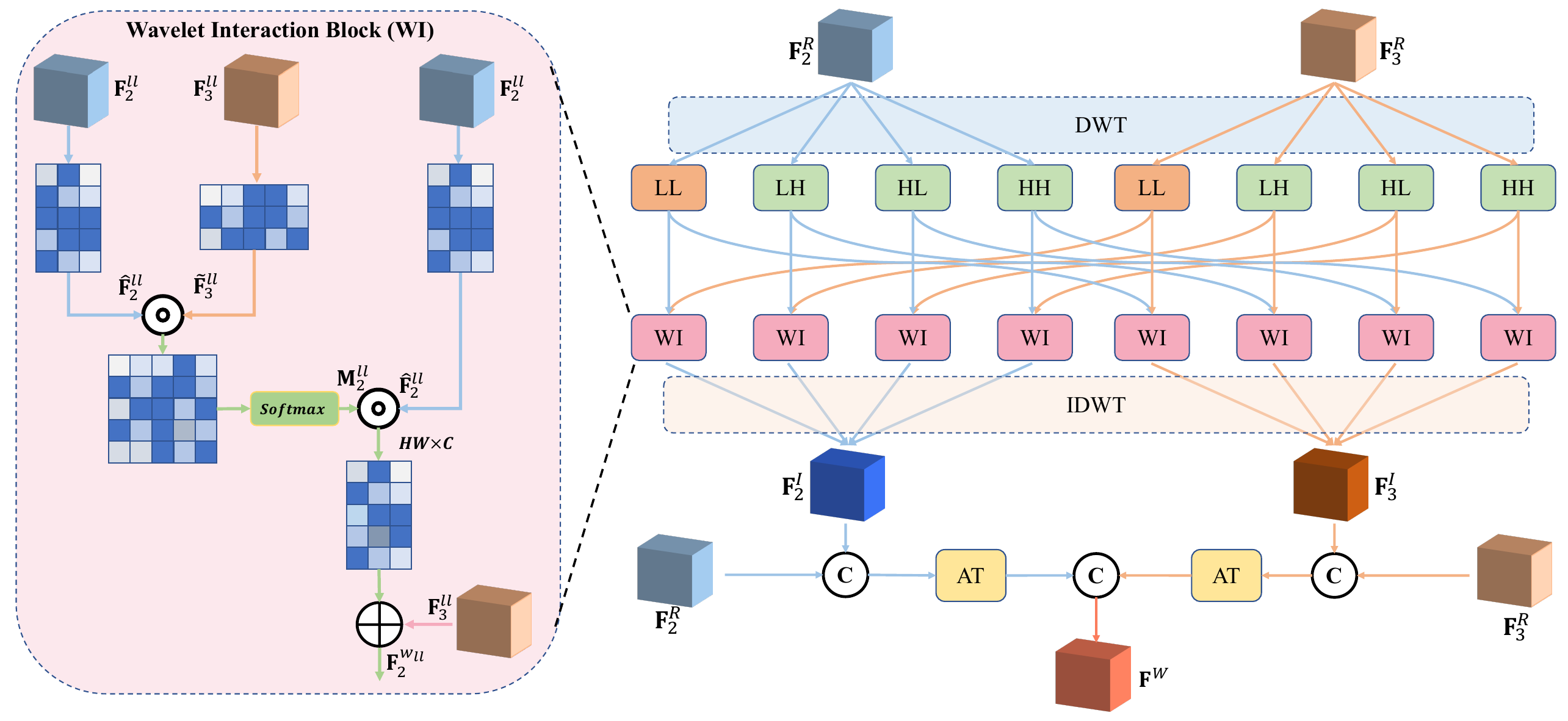}
	\caption{\small{Architecture of the frequency-matching context enhancement  module.}} 
	\label{fig_wie}
\end{figure}
\subsubsection{Wavelet Interaction Stage }
As shown in Fig. \ref{fig_wie}, first,  the discrete wavelet transform is applied to input features $\mathbf{F}_2^{R}$ and $\mathbf{F}_3^{R}$, yielding four different frequency components (\emph{i.e.}, $\mathbf{F}_{2/3}^{ll}$, $\mathbf{F}_{2/3}^{lh}$, $\mathbf{F}_{2/3}^{hl}$, $\mathbf{F}_{2/3}^{hh}$),
\begin{equation}
	\mathbf{F}_{2/3}^{ll}, \mathbf{F}_{2/3}^{lh}, \mathbf{F}_{2/3}^{hl}, \mathbf{F}_{2/3}^{hh} = DWT(\mathbf{F}_{2/3}^R),\\
\end{equation}
where $DWT$ denotes discrete wavelet transform. Then, to achieve the interaction between adjacent layer features, we employ the wavelet interaction block. Take the low-frequency feature components as an example, features $\mathbf{F}_{2}^{ll}$ and  $\mathbf{F}_{3}^{ll}$  $\in \mathbb{R} ^{C\times \frac{H}{2}\times \frac{W}{2}}$ are reshaped to  low dimensional features $\mathbf{\tilde{F}}_2^{ll}$ and $\mathbf{\tilde{F}}_3^{ll}$  $\in R^{C\times \frac{H}{2}\frac{W}{2}}$. We also apply the transpose operation to $\mathbf{\tilde{F}}_2^{ll}$ to transform its dimension into $ \mathbb{R} ^{\frac{H}{2}\frac{W}{2}\times C}$, obtaining $\mathbf{\hat{F}}_2^{ll}$. Next, we employ matrix multiplication and softmax operations on $\mathbf{\hat{F}}_2^{ll}$ and $\mathbf{\tilde{F}}_3^{ll}$ to gain a metric $\mathbf{M}_2^{ll}$ $\in \mathbb{R} ^{\frac{H}{2}\frac{W}{2}\times \frac{H}{2}\frac{W}{2}}$. After that, $\mathbf{M}_2^{ll}$ and $\mathbf{\hat{F}}_2^{ll}$ are subjected to matrix multiplication, transposition
 and reshaping operations. After being added to $\mathbf{F}_{2}^{ll}$, we can obtain feature $\mathbf{F}_2^{w_{ll}}$ which achieves feature interaction and preserves the original feature information . 
\begin{equation}
	\begin{cases}
		\mathbf{\hat{F}}_2^{ll} = (RE(\mathbf{F}_2^{ll}))^T\\
		
		\mathbf{\tilde{F}}_3^{ll} = RE(\mathbf{F}_3^{ll})\\

	\mathbf{M}_2^{ll} = \sigma(\mathbf{\hat{F}}_2^{ll} \odot \mathbf{\tilde{F}}_3^{ll})\\
		
		\mathbf{F}_2^{w_{ll}}  = RE((\mathbf{M}_2^{ll} \odot \mathbf{\hat{F}}_2^{ll})^T)\oplus\mathbf{F}_2^{ll}
		
	\end{cases},
\end{equation}
where  $\odot$ means the matrix multiplication, $RE$ means reshape operation, $\sigma$ denotes the softmax function, and $()^T$ is transposition operation. For other frequency components, we also adopt the above method for feature interaction, yielding $\mathbf{F}_{2/3}^{w_{lh}}$, $\mathbf{F}_{2/3}^{w_{hl}}$ and $\mathbf{F}_{2/3}^{w_{hh}}$. Next, the inverse discrete wavelet transform is leveraged to integrate four frequency components to obtain feature $\mathbf{F}_{2}^{I}$ and $\mathbf{F}_{3}^{I}$$\in \mathbb{R} ^{C\times H\times W}$, 

\begin{equation}
	\mathbf{F}_{2/3}^{I} = IDWT(\mathbf{F}_{2/3}^{w_{ll}},\mathbf{F}_{2/3}^{w_{lh}},\mathbf{F}_{2/3}^{w_{hl}},\mathbf{F}_{2/3}^{w_{hh}}),
\end{equation}
where $IDWT$ means the inverse discrete wavelet transform. Through the above methods, we not only achieve effective feature interaction but also reduce computational complexity by a factor of four.

\subsubsection{Feature Enhancement Stage }

After the wavelet interaction stage, rich contextual information is mined. However, information interaction inevitably introduces other noise information. To this end, we propose the feature enhancement stage to filter out irrelevant information. To be specific, 
feature $	\mathbf{F}_{2}^{I}$ and feature $	\mathbf{F}_{3}^{I}$ are concatenated with input features $	\mathbf{F}_{2}^{R}$ and  $	\mathbf{F}_{3}^{R}$ respectively along the channel dimension. The generated concatenated results go through the AT block  consisting of  channel attention and spatial attention to obtain enhanced features $	\mathbf{F}_{2}^{En}$ and $	\mathbf{F}_{3}^{En}$, which are subjected to concatenation followed by a $3\times3$ convolutional layer for yielding output feature $	\mathbf{F}^{W}$ of FCE module.

\begin{equation}
	\begin{cases}
	\mathbf{F}_{2}^{En} = f_{SA}(f_{CA}(Cat(\mathbf{F}_{2}^{I},\mathbf{F}_{2}^{R})))\\
	\mathbf{F}_{3}^{En} = f_{SA}(f_{CA}(Cat(\mathbf{F}_{3}^{I},\mathbf{F}_{3}^{R})))\\
	\mathbf{F}^{W} = f_{3\times3}(Cat(\mathbf{F}_{2}^{En},\mathbf{F}_{3}^{En}))\\
	\end{cases},
\end{equation}
where $f_{SA}$ and $f_{CA}$ denote spatial attention and channel attention, respectively.

\subsection{Deep Supervision}
To promote the performance of RDNet, we adopt a fusion loss comprising binary cross-entropy (BCE) \cite{de2005tutorial}, boundary intersection over union (IoU) \cite{rahman2016optimizing}, and F-measure (FM) \cite{zhao2019optimizing} losses to supervise saliency prediction. These losses are equally weighted to ensure a balanced optimization objective, as each captures a complementary aspect of segmentation quality, which helps avoid overfitting to any single metric, stabilizes training, and eliminates the need for additional hyperparameter tuning, thereby improving reproducibility. In addition, mean squared error (MSE) loss is employed to supervise the region proportion prediction $\mathbf{F}^{G}$. The overall loss function $L_{\text{total}}$ is defined as follows:
\begin{equation}
	L_{total}=\frac{1}{N} \sum_{i=1}^{N}(L_{bce}+L_{iou}+L_{fm}+L_{mse}),
\end{equation}
where $N$ is batch size of training phase.
\subsubsection{BCE Loss}
\begin{equation}
	L_{bce}=-\sum_{(x,y)}[\mathbf{G}(x,y)log(\mathbf{S}(x,y))+\bar{\mathbf{G}}(x,y)log(\bar{\mathbf{S}}(x,y))] ,
\end{equation}
BCE loss measures the pixel-wise discrepancy between the predicted saliency map and the ground truth, where $\mathbf{S}$ and $\mathbf{G}$ are the predicted result and saliency groundtruth. $\bar{\mathbf{G}}=1-\mathbf{G}$ and  $\bar{\mathbf{S}}=1-\mathbf{S}$.
\subsubsection{IOU Loss}
\begin{equation}
	L_{iou}= 1-\frac{\sum_{(x,y)}\mathbf{S}(x,y)\mathbf{G}(x,y)}{\sum_{(x,y)}[\mathbf{S}(x,y)+\mathbf{G}(x,y)-\mathbf{S}(x,y)\mathbf{G}(x,y)]}.
\end{equation}
IOU loss encourages better region-level overlap between prediction and ground truth by optimizing their intersection-over-union, where $\mathbf{S}(x,y)$ and $\mathbf{G}(x,y)$ are the predicted and ground-truth labels at pixel (x,y), respectively.
\subsubsection{FM Loss}
\begin{equation}
	L_{fm}=1-\frac{(1+\beta^2)TP_{i}}{H_i},
\end{equation}
FM loss evaluates the harmonic mean of precision and recall, balancing them with a weight factor $\beta^2$ to emphasize either precision or recall, where $H=\beta^2(TP+FN)+(TP+FP)$, $TP$, $FP$, and $FN$ denote true positive, false positive, and false negative respectively.
\subsubsection{MSE Loss}
\begin{equation}
L_{mse}=\frac{1}{N} \sum_{i=1}^N (y_{pred}[i] - y_{true}[i])^2
\end{equation}


\begin{figure*}
	\centering
	\includegraphics[width=0.85\linewidth]{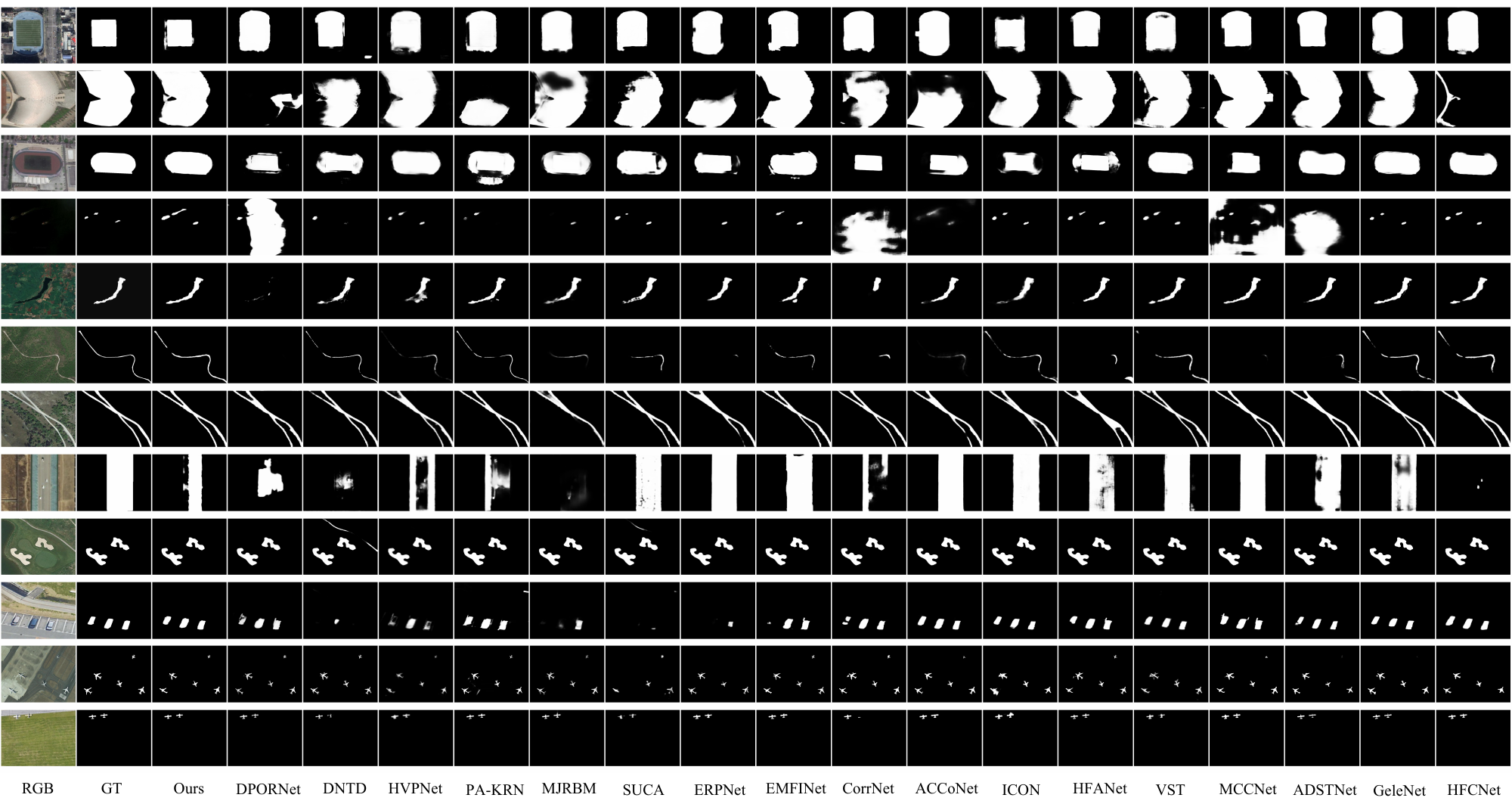}
	\caption{\small{Visual comparison of EORSSD, ORSSD and ORSI-4199 datasets.}} 
	\label{fig_eor}
\end{figure*}



\begin{table*}[t]
	\small
	\renewcommand{\arraystretch}{0.9}
	\setlength\tabcolsep{7pt}
	\centering
	\caption{Quantitative comparisons with 21 methods on EORSSD, ORSSD and ORSI-4199.}
	\begin{tabular}{c|ccc|ccc|ccc}
		
		\hline\hline
		Dataset & \multicolumn{3}{c|}{EORSSD} &\multicolumn{3}{c}{ORSSD}&\multicolumn{3}{c}{ORSI-4199}\\
		\hline
		Metric& $M$$\downarrow$ & $F_{\beta}$$\uparrow$ & $E_{\xi}$  $\uparrow$
		& $M$$\downarrow$ & $F_{\beta}$$\uparrow$ & $E_{\xi}$  $\uparrow$& $M$$\downarrow$ & $F_{\beta}$$\uparrow$ & $E_{\xi}$  $\uparrow$ \\
		\hline
		PoolNet   & 0.0210 & 0.4614 & 0.6853 & 0.0359 & 0.6141 & 0.8142& 0.0543 & 0.7353  & 0.8550 \\
		
		CSNet     & 0.0171 & 0.6284  & 0.8325  & 0.0188 & 0.7574  & 0.9049 & 0.0525 & 0.7151  & 0.8442    \\
		
		R3Net     & 0.0171 & 0.4174  & 0.6472 & 0.0401 & 0.7320  & 0.8713  & 0.0403 & 0.7769 & 0.8828 \\
		
		DPORTNet  & 0.0151 & 0.7377 & 0.8788  & 0.0222 & 0.7861  & 0.8940& 0.0571 & 0.7514  & 0.8641  \\

		DNTD      & 0.0114 & 0.7192  & 0.8714 & 0.0219 & 0.7598  & 0.8862 & 0.0426 & 0.8032  & 0.9019    \\
		
		HVPNet    & 0.0111 & 0.6172  & 0.8098 & 0.0227 & 0.6701  & 0.8485 & 0.0420 & 0.7638  & 0.8845  \\

		PA-KRN    & 0.0106 & 0.7773  & 0.9186  & 0.0141 & 0.8430 & 0.9373& 0.0384 & 0.8161  & 0.9157  \\
		
		MJRBM     & 0.0101 & 0.6994  & 0.8861  & 0.0166 & 0.7951  & 0.9308  & 0.0376 & 0.7975  & 0.9079 \\
		SUCA      & 0.0098 & 0.7161  & 0.8747  & 0.0147 & 0.7693  & 0.9068& 0.0306 & 0.8375  & 0.9256  \\
		
		ERPNet    & 0.0091 & 0.7405  & 0.9153 & 0.0138 & 0.8245  & 0.9489 & 0.0359 & 0.7991  & 0.9025  \\
		EMFINet   & 0.0086 & 0.7752  & 0.9393  & 0.0112 & 0.8461  & 0.9616& 0.0332 & 0.8136 & 0.9123   \\
		CorrNet   & 0.0085 & 0.8092 & 0.9533  & 0.0101 & 0.8720  & 0.9722 & 0.0368 & 0.8489  & 0.9296 \\
		
		ACCoNet   & 0.0076 & 0.7809  & 0.9408   & 0.0092 & 0.8651 & 0.9716& 0.0316 &0.8539 & 0.9393  \\
		ICON      & 0.0074 & 0.7861  & 0.9139  & 0.0119 & 0.8327  & 0.9386  & 0.0284 & 0.8497  & \textcolor{green}{0.9432}\\
		HFANet    & 0.0073 & 0.8082  &0.9243& 0.0096& 0.8668 & 0.9486 & 0.0316 & 0.8271  & 0.9184  \\
		VST       & 0.0069 & 0.7010  & 0.8744 & 0.0096 & 0.8189 & 0.9370& 0.0283 & 0.7929  & 0.9059  \\
		MCCNet    &  0.0068 & 0.7936 & 0.9470 & 0.0091 & 0.8808 &\textcolor{green}{0.9741}& 0.0318 & 0.8550 & 0.9393  \\
		ADSTNet&0.0065&\textcolor{green}{0.8321}&\textcolor{green}{0.9633}&0.0089&\textcolor{green}{0.8856}&\textcolor{blue}{0.9800}&0.0319&\textcolor{green}{0.8615}&0.9412\\
		GeleNet&0.0066&\textcolor{blue}{0.8367}&\textcolor{blue}{0.9678}&\textcolor{green}{0.0083}&\textcolor{blue}{0.8879}&0.9787&0.0266&\textcolor{blue}{0.8711}&\textcolor{blue}{0.9500}\\
		ASTT&\textcolor{green}{0.0059}&0.7534&0.9247&0.0094&0.8456&0.9587&\textcolor{green}{0.0273}&0.8527&0.9216\\
		
		HFCNet    & \textcolor{blue}{0.0051} & 0.7845 & 0.9280& \textcolor{blue}{0.0073} & 0.8581 & 0.9554  &\textcolor{blue}{0.0270}&0.8272&0.9234\\
		Ours       & \textbf{\textcolor{red}{0.0049} }& \textbf{\textcolor{red}{0.8563}}  & \textbf{\textcolor{red}{0.9718}} & \textbf{\textcolor{red}{0.0066} }& \textbf{\textcolor{red}{0.9080}} & \textbf{\textcolor{red}{0.9852}}&\textbf{\textcolor{red}{0.0254} }& \textbf{\textcolor{red}{0.8781}}& \textbf{\textcolor{red}{0.9506} }   \\
		\hline
	\end{tabular}
	\label{tab_comparison}
\end{table*}

\begin{table*}[t]
	\small
	\renewcommand{\arraystretch}{0.9}
	\setlength\tabcolsep{3pt}
	\centering
	\caption{The $t-$test of our method with several compared methods on EORSSD, ORSSD and ORSI-4199 }
	\begin{tabular}{c|ccc|ccc|ccc}
		
		\hline\hline
		Dataset & \multicolumn{3}{c|}{EORSSD} &\multicolumn{3}{c}{ORSSD}&\multicolumn{3}{c}{ORSI-4199}\\
		\hline
		Metric& $P$$
		(M)$ & $P$$(F_{\beta})$ & $P$$(E_{\xi})$  
		& $P$$
		(M)$ & $P$$(F_{\beta})$ & $P$$(E_{\xi})$ & $P$$
		(M)$ & $P$$(F_{\beta})$ & $P$$(E_{\xi})$ \\
		\hline
HFANet& 1.7634$e-$11 & 4.0579$e-$13  & 3.012$e-$13& 3.8254$e-$13 &3.9968$e-$15  &2.3426$e-$14 & 2.6769$e-$15&  1.1102$e-$16 &3.5782$e-$13   \\
VST       & 9.0617$e-$11 & 0  & 5.5511$e-$16& 3.8254$e-$13 & 0 &1.9984$e-$15 &2.506$e-$12 &  0 &   1.9207$e-$14\\
MCCNet    &1.4353$e-$10  &3.6082$e-$14   &9.2075$e-$11 &1.9524$e-$12 & 1.7231$e-$13 & 1.0074$e-$09&2.0111$e-$15 & 1.1313$e-$13  &3.531$e-$09  \\
ADSTNet& 6.6842$e-$10 & 2.2335$e-$10  & 7.4642$e-$07&4.1102$e-$12  & 1.0014$e-$12 &7.2887$e-$07 &1.749$e-$15 &2.1378$e-$12   &1.7031$e-$08 \\
GeleNet& 3.8867$e-$10 &1.5702$e-$09   &1.6749$e-$08& 6.062$e-$11 & 2.6757$e-$12 &1.0895$e-$07 &6.7808$e-$09 &  4.1345$e-$09 &7.2416$e-$2 \\
ASTT& 4.3348$e-$08 & 4.4409$e-$16  &3.2463$e-$13 & 6.062$e-$11 & 1.1102$e-$16 & 4.2577$e-$13&1.1195$e-$10 & 4.8517$e-$14  &9.0716$e-$13 \\
HFCNet    &  7.8857$e-$07&1.0547$e-$14   & 6.1795$e-$13&1.4049$e-$07  &  6.6613$e-$16& 1.4855$e-$13&5.2183$e-$10 & 1.1102$e-$16  &1.6022$e-$12 \\
\hline
\end{tabular}
\label{tab_t}
\end{table*}

\begin{table}[t]
	\centering
	\renewcommand{\arraystretch}{0.95}
	\setlength\tabcolsep{1.3pt}
	\caption{Comparison of the model complexity and the average running speed.}
	\begin{tabular}{c c c c c c c c c}
		\hline
		\hline
		Method&PoolNet&R3Net&PA-KRN&MJRBM&SUCA&ERPNet&EMFINet\\
		Speed(FPS)&80.9&2&16&15.5&24&73.7&25\\
		FLOPs(G)&97.6&47.5&617.7&101.3&56.4&227.4&480.9\\
		\hline
		Method&CorrNet&ACCoNet&HFANet&VST&MCCNet&HFCNet&ADST\\
		Speed(FPS)&100&59.9&26&23&39.5&38&39.5\\
		FLOPs(G)&21.1&368.8&68.3&23.2&234.2&120.41&27.2\\
		\hline
		Method&ASTT&GeleNet&Ours\\
		Speed(FPS)&13&25.45&13.6\\
		FLOPs(G)&69.5&11.7&48.7\\
		\hline
	\end{tabular}
	\label{tab_time}
\end{table}
\section{Experiments And Analyses}
\subsection{Implementation Details and Evaluation Metrics}
\subsubsection{Implementation Details}
We train and test our RDNet on three public datasets ORSSD \cite{li2019nested}, EORSSD \cite{zhang2020dense}, and ORSI-4199 \cite{tu2021orsi}. The ORSSD dataset, the first public dataset for ORSI-SOD, includes 800 images with corresponding pixel-level ground truths (GTs), divided into 600 images for training and 200 for testing. The EORSSD dataset comprises 2,000 images with GTs, split into 1,400 for training and 600 for testing. The ORSI-4199 dataset, the largest dataset for ORSI-SOD, contains 4,199 images with GTs, with 2,000 images used for training and 2,199 for testing. We implement our RDNet in the PyTorch framework and leverage the workstation with a single NVIDIA RTX 3090 to accelerate the training process. For the backbone network, we adopt SwinTransformer as the backbone, and initialize it with the pre-trained parameters. During the training process, we resize all training images to $384\times 384$ and apply data augmentation techniques such as random flipping, clipping, and rotating. To optimize the training process, we use the RMSprop optimizer \cite{tieleman2012rmsprop} to minimize the loss function. We set the initial learning rate to $1e-5$, the momentum to 0.9 and the batch size to 4.

\subsubsection{Evaluation Metrics}
To assess the performance of the proposed RDNet, we adopt three commonly used evaluation metrics: mean absolute error ($M$) \cite{perazzi2012saliency}, F-measure ($F_{\beta}$) \cite{achanta2009frequency}, and  E-measure ($E_{\xi}$) \cite{fan2018enhanced}. The mathematical definitions of these metrics are provided below:

\emph{(a)}:
\begin{equation}
	M=\frac{1}{W\times H}{\textstyle \sum_{x=1}^{W}}{\textstyle \sum_{y=1}^{H}}|\mathbf{S}(x,y)-\mathbf{G}(x,y)|,
\end{equation}
where $\mathbf{S}(x,y)$ and $\mathbf{G}(x,y)$ mean predicted saliency map and groundtruth respectively.

\emph{(b)}:
\begin{equation}
	F_{\beta}=\frac{(1+\beta^2)Precision\cdot Recall}{\beta^2\cdot Precision+Recall},
\end{equation}
where $Precision$ and $Recall$ values are used for yielding the PR curve.

\emph{(c)}:
\begin{equation}
	E_{\xi}=\frac{1}{W\times H}{\textstyle \sum_{x=1}^{W}}{\textstyle \sum_{y=1}^{H}}\theta(x,y),
\end{equation}
where $\theta$ means the relation between $\mathbf{S}$ and $\mathbf{G}$.


\subsection{Comparison with State-of-the-Art Methods}
In experiments, we compare the proposed RDNet with other state-of-the-art NSI-SOD and ORSI-SOD methods, including R3Net \cite{deng2018r3net}, PoolNet \cite{liu2019simple}, CSNet \cite{gao2020highly}, SUCA \cite{li2020stacked}, PA-KRN \cite{xu2021locate}, VST \cite{liu2021visual}, DPORTNet \cite{liu2022disentangled}, DNTD \cite{fang2022densely}, ICON \cite{zhuge2022salient}, MJRBM \cite{tu2021orsi}, EMFINet \cite{wang2022multiscale}, ERPNet \cite{zhou2022edge}, ACCoNet \cite{li2022adjacent}, CorrNet \cite{li2023lightweight}, MCCNet \cite{li2021multi}, ADST \cite{zhao2024adaptive}, ASTT \cite{gao2023adaptive}, GeleNet \cite{li2023salient}, HFANet \cite{wang2022hybrid}, and HFCNet \cite{liu2024heterogeneous}.  To ensure a fair comparison, we used the source code released online by the  authors.

\subsubsection{Qualitative Comparison}
To demonstrate the effectiveness  of our  RDNet, we present some visualization results of challenging scenes (\emph{i.e.}, big salient object (BSO), small salient object (SSO), narrow salient object (NAO), multiple salient object (MSO)) from EORSSD \cite{zhang2020dense}, ORSSD \cite{li2019nested} and ORSI-4199 \cite{tu2021orsi} in Fig. \ref{fig_eor}.
\paragraph{Advantages in Big Salient Object}
In salient object detection, scenes with large objects often pose difficulties for detection. For example, in the $1^{st}$ row and  $3^{rd}$ row of Fig. \ref{fig_eor}, the stadium occupies a significant portion of the images. From the detection results, it is clear that most methods struggle to accurately capture edge details to predict the salient objects in their entirety. In addition, in the $2^{rd}$ row  of Fig. \ref{fig_eor}, it can be seen that most methods (\emph{i.e.}, DPORNet, PA-KRN and HFCNet) fail to fully restore the complete object region .This limitation is due to the challenges these methods encounter in effectively capturing shallow detail features.

\paragraph{Advantages in Narrow Salient Object}

Due to the relatively high shooting altitude of remote sensing images, the objects often appear extremely narrow in the pictures, which poses great difficulties for detection. As shown in Fig. \ref{fig_eor} ($5^{th}$, $6^{th}$, $7^{th}$ and $8^{th}$ rows), we list some three narrow salient object from three datasets. For example, in the $6^{th}$ row of the figure, a long and narrow river is depicted. From the visualized detection results, it is evident that multiple methods (\emph{i.e.}, HVPNet, MJRBM, ACCoNet) fail to fully capture and reconstruct the complete structure of the river. In addition, in the $8^{th}$ of Fig. \ref{fig_eor} , there is a narrow road stretches across the entire image. Form it, we can see that MJRBM and HFCNet fail to detect some regions, while HVPNet, CorrNet and HFANet manage to detect only a limited portion of these areas. The reason for this phenomenon is that these methods lack the capability to extract global features to capture the overall shape and structure of the object.

\paragraph{Advantages in Multiple Salient Object}
Multi-object detection has consistently been a significant challenge in saliency detection. For example, in the $4^{th}$ row, where three boats sail in the sea , we can see that several methods (\emph{i.e.}, DNTD, MJRBM and EPRNet ) cannot integrally recover the object and some methods misregard the background (\emph{i.e.}, DPORNet, CorrNet and MCCNet) as the saliency region. Similarly, in the $10^{th}$ row of  Fig. \ref{fig_eor}, three cars (prominent objects) are parked side by side in the image, from the comparison results, we can observe that method DNTD, MJRBM and ERPNet are only able to identify one or two cars in the image.

\paragraph{Advantages in Small Salient Object}

Detecting small salient objects is difficult due to scarce details and low spatial resolution. For example, in the the $4^{th}$, $11^{th}$ and $12^{th}$ rows, due to the shooting altitude, the boats and airplanes in the image appear extremely small, making it difficult to discern their details even with the naked eye. Through analysis of the experimental results, we can observe that most methods lack the ability to recover fine details of small objects.

By analyzing the visual results of multiple comparative methods, it is evident that our approach consistently achieves superior detection performance across various complex scenarios. This advantage primarily stems from a key limitation in most existing methods, which tend to adopt a unified feature extraction and optimization strategy regardless of object scale. Such an approach overlooks the inherent differences in structure, semantics, and contextual dependencies among objects of varying sizes. As a result, small objects are often missed, while large objects may suffer from blurred boundaries or incomplete segmentation. In contrast, our method incorporates a region-proportion-aware strategy, which effectively addresses these issues and significantly enhances the model's generalization ability and detection accuracy across multi-scale scenes.

\begin{figure}[!t]
	\centering
	\includegraphics[width=0.45\textwidth]{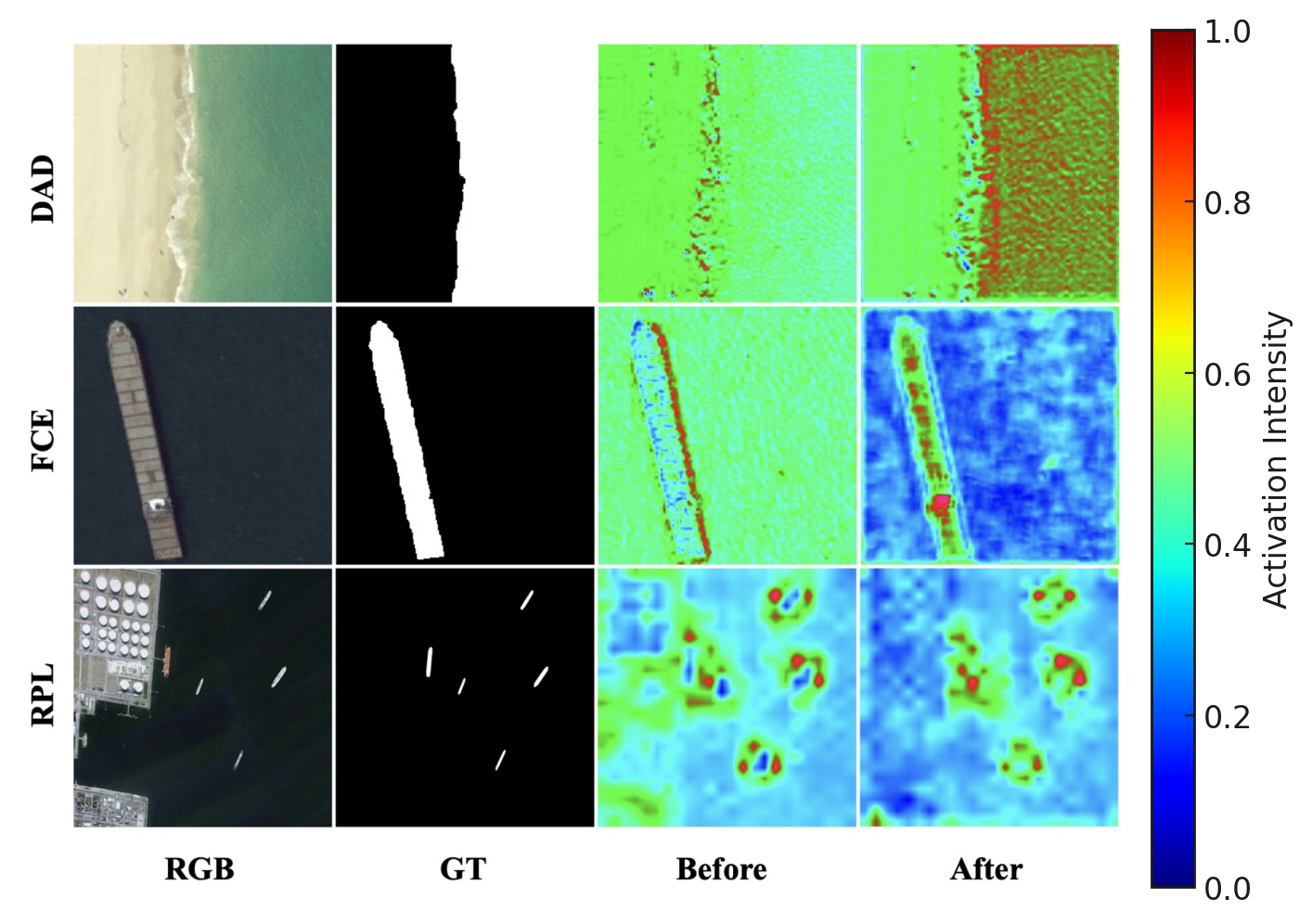}
	\caption{\small{Illustration of the impact of different module.}} 
	\label{fig_hot}
\end{figure}

\begin{figure}[!t]
	\centering
	\includegraphics[width=0.43\textwidth]{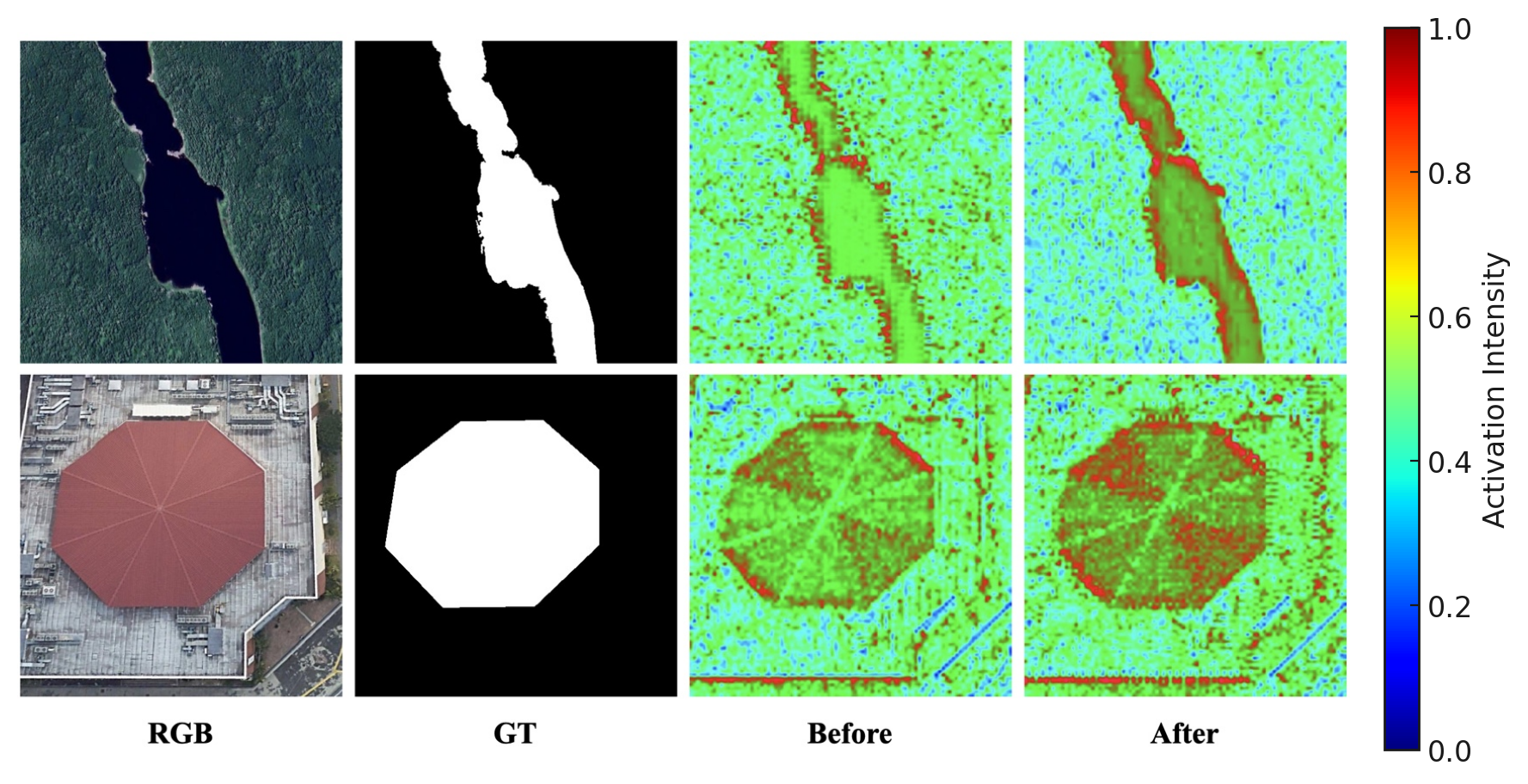}
	\caption{\small{Illustration of the impact of wavelet interaction.}} 
	\label{fig_wi}
\end{figure}

\begin{table}[t]
\footnotesize
	\renewcommand{\arraystretch}{0.95}
	\setlength\tabcolsep{8pt}
	\centering
	\caption{Ablation study on different modules.}
	\begin{tabular}{cccc}
		\hline
		\multicolumn{1}{c}{Settings}&  \multicolumn{1}{c}{$M$$\downarrow$}&\multicolumn{1}{c}{$F_{\beta}$$\uparrow$}&  \multicolumn{1}{c}{$S_{\alpha}$$\uparrow$}\\
		\hline
		w/o DAD&0.0052&0.8550&0.9273\\
		w/o FCE&0.0061&0.8453&0.9224\\
		w/o RPL&0.0054&0.8561&0.9329\\
		\textbf{Ours}&\textbf{0.0049}&\textbf{0.8563}&\textbf{0.9327}\\
		\hline
		
		\hline
	\end{tabular}
	\label{tab_wo}
\end{table}

\begin{table}[t]
	\footnotesize
	\renewcommand{\arraystretch}{0.95}
	\setlength\tabcolsep{8pt}
	\centering
	\caption{Ablation study on DAD module.}
	\begin{tabular}{ccccc}
		\hline
		\multicolumn{1}{c}{Settings}&  \multicolumn{1}{c}{$M$$\downarrow$}&\multicolumn{1}{c}{$F_{\beta}$$\uparrow$}& \multicolumn{1}{c}{$S_{\alpha}$$\uparrow$}\\
		\hline
		DAD$\_$1&0.0051&0.8494&0.9316\\
		DAD$\_$2&0.0051&0.8540&0.9322\\
		DAD$\_$3&0.0054&0.8589&0.9314\\
		w/o DO&0.0059&0.8505&0.9283\\
		w/o DE&0.0053&0.8502&0.9256\\
		\textbf{Ours}&\textbf{0.0049}&\textbf{0.8563}&\textbf{0.9327}\\
		\hline
		
		\hline
	\end{tabular}
	\label{tab_PMD}
\end{table}

\begin{table}[t]
\footnotesize
	\renewcommand{\arraystretch}{0.95}
	\setlength\tabcolsep{8pt}
	\centering
	\caption{Ablation study on FCE modules.}
	\begin{tabular}{cccc}
		\hline
		\multicolumn{1}{c}{Settings}&  \multicolumn{1}{c}{$M$$\downarrow$}&\multicolumn{1}{c}{$F_{\beta}$$\uparrow$}&  \multicolumn{1}{c}{$S_{\alpha}$$\uparrow$}\\
		\hline
		w/o WI&0.0059&0.8562&0.9271\\
		w/o FE&0.0057&0.8511&0.9270\\
		\textbf{Ours}&\textbf{0.0049}&\textbf{0.8563}&\textbf{0.9327}\\
		\hline
		
		\hline
	\end{tabular}
	\label{tab_WIE}
\end{table}

\begin{table}[t]
\footnotesize
	\renewcommand{\arraystretch}{0.95}
	\setlength\tabcolsep{6pt}
	\centering
	\caption{Ablation study on different backbones.}
	\begin{tabular}{ccccc}
		\hline
		\multicolumn{1}{c}{Settings}&  \multicolumn{1}{c}{$M$$\downarrow$}&\multicolumn{1}{c}{$F_{\beta}$$\uparrow$}& \multicolumn{1}{c}{$S_{\alpha}$$\uparrow$}\\
		\hline
		
		ResNet&0.0094&0.7662&0.9127\\
		VGG&0.0134&0.7379&0.8680\\
		PVT&0.0073&0.8194&0.9214\\
		ViT&0.0175&0.5742&0.7832\\
		\textbf{Ours}&\textbf{0.0049}&\textbf{0.8563}&\textbf{0.9327}\\
		\hline
		
		\hline
	\end{tabular}
	\label{tab_BB}
\end{table}

\begin{table}[t]
\footnotesize
	\renewcommand{\arraystretch}{0.95}
	\setlength\tabcolsep{6pt}
	\centering
	\caption{Ablation study on different region proportion estimation methods.}
	\begin{tabular}{ccccc}
		\hline
		\multicolumn{1}{c}{Settings}&  \multicolumn{1}{c}{$M$$\downarrow$}&\multicolumn{1}{c}{$F_{\beta}$$\uparrow$}& \multicolumn{1}{c}{$S_{\alpha}$$\uparrow$}\\
		\hline
		
		Tradition&0.0056&0.8553&0.9283\\
		Attention&0.0050&0.8560&0.9328\\
		\textbf{Ours}&\textbf{0.0049}&\textbf{0.8563}&\textbf{0.9327}\\
		\hline
		
		\hline
	\end{tabular}
	\label{tab_R}
\end{table}

\begin{table}[t]
\footnotesize
	\renewcommand{\arraystretch}{0.95}
	\setlength\tabcolsep{6pt}
	\centering
	\caption{Ablation study on different threshold settings.}
	\begin{tabular}{ccccc}
		\hline
		\multicolumn{1}{c}{Settings}&  \multicolumn{1}{c}{$M$$\downarrow$}&\multicolumn{1}{c}{$F_{\beta}$$\uparrow$}& \multicolumn{1}{c}{$S_{\alpha}$$\uparrow$}\\
		\hline
		
		T1&0.0052&0.8553&0.9304\\
		T2&0.0055&0.8550&0.9299\\
		\textbf{Ours}&\textbf{0.0049}&\textbf{0.8563}&\textbf{0.9327}\\
		\hline
		
		\hline
	\end{tabular}
	\label{tab_T}
\end{table}

\begin{table}[t]
\footnotesize
	\renewcommand{\arraystretch}{0.95}
	\setlength\tabcolsep{6pt}
	\centering
	\caption{Ablation study on different feature interaction method.}
	\begin{tabular}{ccccc}
		\hline
		\multicolumn{1}{c}{Settings}&  \multicolumn{1}{c}{$M$$\downarrow$}&\multicolumn{1}{c}{$F_{\beta}$$\uparrow$}& \multicolumn{1}{c}{$S_{\alpha}$$\uparrow$}\\
		\hline
		
		SE&0.0054&0.8525&0.9300\\
		CBAM&0.0053&0.8552&0.9306\\
		Addition&0.0054&0.8561&0.9329\\
		\textbf{Ours}&\textbf{0.0049}&\textbf{0.8563}&\textbf{0.9327}\\
		\hline
		
		\hline
	\end{tabular}
	\label{tab_FI}
\end{table}

\subsubsection{Quantitative comparison}  

In this section, we present a quantitative performance comparison of our RDNet with other SOD methods using three evaluation metrics, as detailed in Table \ref{tab_comparison}. The results demonstrate that our proposed method outperforms the others across all three datasets. 

Table. \ref{tab_comparison}  presents the quantitative comparison of our method with 21 other methods on the EORSSD, ORSSD and ORSI-4199 datasets, which show that our RDNet outperforms all the compared methods on both datasets.To be specific, on the EORSSD dataset, for the $M$, where smaller is better, our method yields the smallest result (0.0049) which is $3.9\%$ lower than the best method HFCNet\cite{liu2024heterogeneous}. For the other metrics, our method also gain the best result, which achieves an average improvement of $9.1\%$ and $4.7\%$   in $F_{\beta}$ and $E_{\xi}$. Compared to the transformer-based method ASTT \cite{gao2023adaptive}, our method has achieved significant performance improvements in three metrics, with an increase of 13.6$\%$  in $F_{\beta}$   and 5.1$\%$ in $E_{\xi}$, and a decrease of 16.9$\%$  in $M$. On the ORSSD dataset, our RDNet achieve the best results in three metrics and  notably surpasses the  method ADSTNet \cite{zhao2024adaptive} by $2.5\%$ and $0.5\%$  in terms of $F_{\beta}$ and $E_{\xi}$. Besides, Table. \ref{tab_comparison} provides the quantitative comparison of our RDNet with 21 other methods  on the ORSI-4199 dataset. Concretely, our method is able to bring $10.2\%$, $10.7\%$ and $4.9\%$ in $M$, $F_{\beta}$ and $E_{\xi}$ compared to VST  \cite{liu2021visual}. In addition, to demonstrate the performance improvement of our method more clearly, we conduct t-tests comparing our proposed method with the six best-performing baseline methods. As shown in Table. \ref{tab_t}, the p-values for each baseline method across the three datasets are all relatively small, indicating that our improvements are statistically significant.  Furthermore, we list the average running speed and model complexity  of several detection methods. As shown in Table. \ref{tab_time}, compared with other learning-based methods, RDNet has a relatively small model complexity. For the average running speed, considering that our method involves extensive matrix operations, it achieves  13 FPS.

\subsection{Ablation Study}
 
To show the effectiveness of each component in our RDNet (\emph{i.e.}, DAD, FCE and RPL), we conduct several ablation experiments, all numerical results are presented in  Table. \ref{tab_wo}.
\subsubsection{Effectiveness of different modules}
To evaluate the effectiveness of the modules proposed in this paper, we conduct three separate experiments, each involving the removal of one module. For instance, in experiment ``$w/o$ FCE", FCE module is  replaced with a simple addition operation. From  the numerical results of Table. \ref{tab_wo}, we can observe that the model's performance decreases as modules are removed. Notably, compared to the ``$w/o$ DAD", complete model (Ours) has decrease of $5.7\%$ in terms of MAE and improvement of $0.2\%$  in terms of $F_{\beta}$. Furthermore, we visualize the feature maps before and after each module. For example, for the 1$^{st}$ row in Fig. \ref{fig_hot}, we can observe that before applying the DAD module, the model primarily focuses on the boundary between the beach and the sea. After incorporating the DAD module, however, the attention shifts towards the sea region, demonstrating that the DAD module effectively enhances the detection of fine-grained target details. Besides, in the 3$^{rd}$ row in Fig. \ref{fig_hot}, which contains five vessels, a comparison of the feature maps before and after the RPL module reveals that the RPL module leads to more focused activations on the target regions.

\subsubsection{Effectiveness of each component in DAD module}
First, to verify the effectiveness of proportion guidance, we remove the PG block, ``DAD\_1", ``DAD\_2", ``DAD\_3" represent the three detailed information extraction strategies shown in Fig. \ref{fig_large}. From Table. \ref{tab_PMD}, we can conclude that the use of the proportion guidance enables the detection network to accurately extract detailed information and significantly improve the detection performance. Secondly, to illustrate the effectiveness  of two branches of DAD module, we conduct two experiments. To be specific, for the ``$w/o$ DE", we delete all the convolutional layers in the detail extractor (Lower branch), and perform multiplication and addition operations on the weights obtained from the upper branch and the input features, where generated features serve as the output of the DAD module. For the ``$w/o$ DO", we remove the upper branch and add the output features of all convolutional layers in the Lower branch as the output of the DAD module. From all numerical results  in Table. \ref{tab_PMD}, it is evident that convolutional layers of different sizes and the attention mechanism are crucial for improving the detection performance.
\subsubsection{Effectiveness of each component in FCE module}
To illustrate the effectiveness of the two stages in the WIE module, we conduct two experiments presented in Table. \ref{tab_WIE}. For the ``$w/o$ WI", we remove the wavelet interaction stage and directly feed the two input features into the feature enhancement stage for optimization and fusion. Besides, to intuitively demonstrate the effectiveness of  the wavelet interaction, we provide the feature maps in Fig. \ref{fig_wi}, it can be observed that the wavelet interaction significantly enhances both the boundary clarity and the integrity of the  object.
In addition, for the ``$w/o$ FE", the feature enhancement stage is deleted, two output features of the wavelet interaction stage are directly added together to obtain the output feature of FCE module. From the Table. \ref{tab_WIE}, numerical results reveal the importance of two stage. 

\subsubsection{Effectiveness of different backbone network}

To verify the impact of different backbone networks on  RDNet, we replace the SwinTransformer with VGG-16\cite{simonyan2014very} and ResNet-34\cite{he2016deep} as shown in Table. \ref{tab_BB}. Compared to ResNet-34, our model gains improvement of $11.7\%$  in terms of $F_{\beta}$. Besides,  compared to VGG-16, our RDNet obtains $16.0\%$ and $7.4\%$ in terms of $F_{\beta}$ and $S_{\alpha}$, respectively. This suggests that SwinTransformer has a stronger feature extraction capability than VGG-16 and ResNet-34. Moreover, we also replace the SwinTransformer with ViT \cite{dosovitskiy2020image} and PVT \cite{wang2021pyramid}, as shown in Table. \ref{tab_BB} (3$^{rd}$ and 4$^{th}$ rows). To be specific, compared to ViT, our method achieves a significant performance improvement, with a 72$\%$ reduction in $M$ and a 49$\%$ increase in $F_{\beta}$. Compared to PVT, our model also  obtains $4.5\%$ and $1.2\%$ in terms of $F_{\beta}$ and $S_{\alpha}$.The reason behind the above performance improvement is that Swin Transformer outperforms ViT and PVT due to its hierarchical design and shifted window attention, which effectively capture both local details and global context. Unlike ViT's flat structure or PVT's costly global attention, Swin Transformer achieves better efficiency and multi-scale feature representation, making it more suitable for dense prediction tasks.

\subsubsection{Effectiveness of different region proportion estimation methods}

To verify the impact of different region proportion estimation methods, we conduct two more experiments, as shown in Table. \ref{tab_R}, where the our method (global average pooling and FC layers) is replaced with tradition-based and attention-based methods. For the tradition-based method, we use Sobel gradient filtering and thresholding to estimate the region proportion, from the experimental results, we observe that incorporating the traditional method actually leads to a decline in detection performance, indicating that it fails to accurately estimate the target region proportion. Besides, for the attention-based method, we performed spatial attention on the features before applying global average pooling. However, the comparative results shown in Table. \ref{tab_R}, illustrating that incorporating the attention mechanism did not lead to any noticeable improvement in model performance.

\subsubsection{Effectiveness of different threshold settings}
To validate the effectiveness of the threshold setting adopted in this work, we conducted two additional experiments with alternative threshold configurations, and the results are reported in Table 1. Specifically, T1 represents ``[50$\%$$<$, 50$\%$$-$75$\%$, $>$75$\%$]" and T2 represents ``[10$\%$$<$, 10$\%$$-$30$\%$, $>$30$\%$]". From the comparisons, it is evident that altering the threshold setting leads to performance degradation. This can be attributed to the fact that when the bin ranges are too broad, large objects may be processed with small kernels, limiting attention to only a portion of the object. Conversely, when the bin ranges are too narrow, large kernels may be applied to small objects, causing the convolution to incorporate excessive background information, thereby compromising feature representation.

\subsubsection{Effectiveness of different feature interaction methods}
To verify the effectiveness of the feature interaction strategy adopted in the RPL module, we conducted a comparison using three alternative lightweight interaction methods, as shown in Table. \ref{tab_FI}. For the ``SE", the concatenation result of $\mathbf{F}_4^{R} $ and feature $\mathbf{F}_5^{R}$ is fed into the Squeeze-and-Excitation block. Similarly, for the ``CBAM", the concatenation result of $\mathbf{F}_4^{R} $ and feature $\mathbf{F}_5^{R}$ is fed into the Convolutional Block Attention block. In addition, we also replace the fusion method with addition operation. From the comparative results in the Table. \ref{tab_FI}, it can be seen that such simplified feature interaction methods may fall short in capturing complex object information in challenging scenes and cannot replace the proposed RPL module in effectively modeling channel and spatial interactions.
\begin{figure}[!t]
	\centering
	\includegraphics[width=0.4\textwidth]{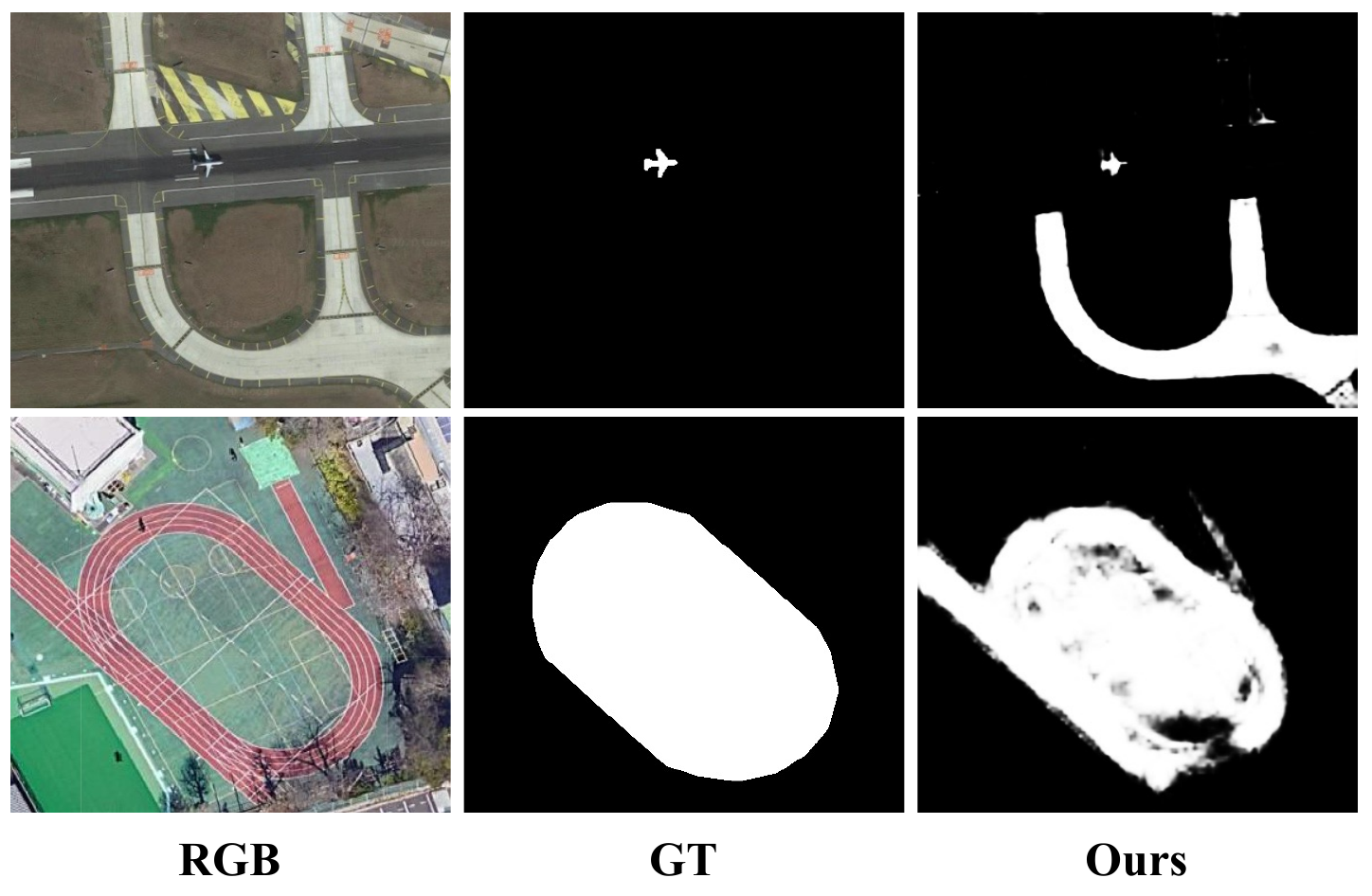}
	\caption{\small{Example of failure case.}} 
	\label{fig_failure}
\end{figure}
\subsection{Failure cases}
To give a comprehensive presentation of our RDNet, we list some failure cases. As shown in Fig. \ref{fig_failure} (1$^{st}$ row), when the salient objects in an image are extremely small or fine, our network struggles to detect them accurately and also detects the runway surrounding the airplane as a salient object. for the second row in Fig. \ref{fig_failure} , when the background shares similar textures with the target, our method also identifies the extended runway from the playground as part of the target region.
\section{Conclusion}
This paper presents the Region Proportion-aware Dynamic Adaptive Salient Object Detection Network (RDNet) for ORSI-SOD, which replaces traditional CNNs with the SwinTransformer to better capture global context. RDNet integrates three key modules: the dynamic adaptive detail-aware (DAD) module, which leverages proportion guidance to adaptively apply varied convolution kernels for multi-scale feature extraction; the frequency-matching context enhancement (FCE) module, which enriches contextual information through wavelet interactions and refines features with channel and spatial attention mechanisms; and the region proportion-aware localization (RPL) module, which employs continuous cross-attention on high-level features to focus on semantic information and improve the detection of randomly located objects. The main objectives of this work are to enhance robustness against large-scale variations and to achieve precise localization of salient objects in complex optical remote sensing scenarios. Extensive experiments demonstrate that RDNet effectively achieves these goals and delivers superior detection performance compared with state-of-the-art methods.
\bibliographystyle{IEEEtran}
\bibliography{reference}

@article{de2005tutorial,
	title={A tutorial on the cross-entropy method},
	author={De Boer, Pieter-Tjerk and Kroese, Dirk P and Mannor, Shie and Rubinstein, Reuven Y},
	journal={Annals of operations research},
	volume={134},
	number={1},
	pages={19--67},
	year={2005},
	publisher={Springer}
}

@inproceedings{rahman2016optimizing,
	title={Optimizing intersection-over-union in deep neural networks for image segmentation},
	author={Rahman, Md Atiqur and Wang, Yang},
	booktitle={International symposium on visual computing},
	pages={234--244},
	year={2016},
	organization={Springer}
}

@inproceedings{zhao2019optimizing,
	title={Optimizing the f-measure for threshold-free salient object detection},
	author={Zhao, Kai and Gao, Shanghua and Wang, Wenguan and Cheng, Ming-Ming},
	booktitle={Proceedings of the IEEE/CVF International Conference on Computer Vision},
	pages={8849--8857},
	year={2019}
}

@inproceedings{perazzi2012saliency,
	title={Saliency filters: Contrast based filtering for salient region detection},
	author={Perazzi, Federico and Kr{\"a}henb{\"u}hl, Philipp and Pritch, Yael and Hornung, Alexander},
	booktitle={2012 IEEE Conference on Computer Vision and Pattern Recognition},
	pages={733--740},
	year={2012},
	
}

@inproceedings{achanta2009frequency,
	title={Frequency-tuned salient region detection},
	author={Achanta, Radhakrishna and Hemami, Sheila and Estrada, Francisco and Susstrunk, Sabine},
	booktitle={2009 IEEE Conference on Computer Vision and Pattern Recognition},
	pages={1597--1604},
	year={2009},
	
}

@article{fan2018enhanced,
	title={Enhanced-alignment measure for binary foreground map evaluation},
	author={Fan, Deng-Ping and Gong, Cheng and Cao, Yang and Ren, Bo and Cheng, Ming-Ming and Borji, Ali},
	journal={arXiv preprint arXiv:1805.10421},
	year={2018}
}

@article{simonyan2014very,
	title={Very deep convolutional networks for large-scale image recognition},
	author={Simonyan, Karen and Zisserman, Andrew},
	journal={arXiv preprint arXiv:1409.1556},
	year={2014}
}

@inproceedings{he2016deep,
	title={Deep residual learning for image recognition},
	author={He, Kaiming and Zhang, Xiangyu and Ren, Shaoqing and Sun, Jian},
	booktitle={Proceedings of the IEEE Conference on Computer Vision and Pattern Recognition},
	pages={770--778},
	year={2016}
}

@article{tieleman2012rmsprop,
	title={Rmsprop: Divide the gradient by a running average of its recent magnitude. coursera: Neural networks for machine learning},
	author={Tieleman, Tijmen and Hinton, Geoffrey},
	journal={COURSERA Neural Networks Mach. Learn},
	year={2012}
}

@article{piao2021panet,
	title={PANet: Patch-aware network for light field salient object detection},
	author={Piao, Yongri and Jiang, Yongyao and Zhang, Miao and Wang, Jian and Lu, Huchuan},
	journal={IEEE Transactions on Cybernetics},
	year={2021},
	publisher={IEEE}
}

@article{zhuge2022cubenet,
	title={CubeNet: X-shape connection for camouflaged object detection},
	author={Zhuge, Mingchen and Lu, Xiankai and Guo, Yiyou and Cai, Zhihua and Chen, Shuhan},
	journal={Pattern Recognition},
	volume={127},
	pages={108644},
	year={2022},
	publisher={Elsevier}
}

@article{chen2022camouflaged,
	title={Camouflaged object detection via context-aware cross-level fusion},
	author={Chen, Geng and Liu, Si-Jie and Sun, Yu-Jia and Ji, Ge-Peng and Wu, Ya-Feng and Zhou, Tao},
	journal={IEEE Transactions on Circuits and Systems for Video Technology},
	volume={32},
	number={10},
	pages={6981--6993},
	year={2022},
	publisher={IEEE}
}

@article{huang2022small,
	title={Small object detection method with shallow feature fusion network for chip surface defect detection},
	author={Huang, Haixin and Tang, Xueduo and Wen, Feng and Jin, Xin},
	journal={Scientific reports},
	volume={12},
	number={1},
	pages={3914},
	year={2022},
	publisher={Nature Publishing Group UK London}
}

@article{zhang2020dense,
	title={Dense attention fluid network for salient object detection in optical remote sensing images},
	author={Zhang, Qijian and Cong, Runmin and Li, Chongyi and Cheng, Ming-Ming and Fang, Yuming and Cao, Xiaochun and Zhao, Yao and Kwong, Sam},
	journal={IEEE Transactions on Image Processing},
	volume={30},
	pages={1305--1317},
	year={2020},
	publisher={IEEE}
}

@article{zhou2022edge,
	title={Edge-guided recurrent positioning network for salient object detection in optical remote sensing images},
	author={Zhou, Xiaofei and Shen, Kunye and Weng, Li and Cong, Runmin and Zheng, Bolun and Zhang, Jiyong and Yan, Chenggang},
	journal={IEEE Transactions on Cybernetics},
	year={2022},
	note={doi:{10.1109/TCYB.2022.3163152}},
	publisher={IEEE}
}

@article{wan2023lfrnet,
	title={LFRNet: Localizing, Focus, and Refinement Network for Salient Object Detection of Surface Defects},
	author={Wan, Bin and Zhou, Xiaofei and Zheng, Bolun and Yin, Haibing and Zhu, Zunjie and Wang, Hongkui and Sun, Yaoqi and Zhang, Jiyong and Yan, Chenggang},
	journal={IEEE Transactions on Instrumentation and Measurement},
	year={2023},
	note={doi:{10.1109/TIM.2023.3250302}},
	publisher={IEEE}
}

@article{zhang2020lfnet,
	title={LFNet: Light field fusion network for salient object detection},
	author={Zhang, Miao and Ji, Wei and Piao, Yongri and Li, Jingjing and Zhang, Yu and Xu, Shuang and Lu, Huchuan},
	journal={IEEE Transactions on Image Processing},
	volume={29},
	pages={6276--6287},
	year={2020},
	publisher={IEEE}
}

@article{li2023lightweight,
	title={Lightweight salient object detection in optical remote-sensing images via semantic matching and edge alignment},
	author={Li, Gongyang and Liu, Zhi and Zhang, Xinpeng and Lin, Weisi},
	journal={IEEE Transactions on Geoscience and Remote Sensing},
	volume={61},
	pages={1--11},
	year={2023},
	publisher={IEEE}
}

@article{zeng2023adaptive,
	title={Adaptive edge-aware semantic interaction network for salient object detection in optical remote sensing images},
	author={Zeng, Xiangyu and Xu, Mingzhu and Hu, Yijun and Tang, Haoyu and Hu, Yupeng and Nie, Liqiang},
	journal={IEEE Transactions on Geoscience and Remote Sensing},
	year={2023},
	publisher={IEEE}
}

@article{li2023salient,
	title={Salient object detection in optical remote sensing images driven by transformer},
	author={Li, Gongyang and Bai, Zhen and Liu, Zhi and Zhang, Xinpeng and Ling, Haibin},
	journal={IEEE Transactions on Image Processing},
	year={2023},
	publisher={IEEE}
}

@article{gongyangli2022lightweight,
	title={Lightweight salient object detection in optical remote sensing images via feature correlation},
	author={GongyangLi, Z and Bai, Zhen and Lin, Weisi and Ling, Haibin},
	journal={IEEE Trans. Geosci. Remote Sens.},
	volume={60},
	year={2022}
}

@article{vaswani2017attention,
	title={Attention is all you need},
	author={Vaswani, A},
	journal={Advances in Neural Information Processing Systems},
	year={2017}
}

@inproceedings{deng2018r3net,
	title={R3net: Recurrent residual refinement network for saliency detection},
	author={Deng, Zijun and Hu, Xiaowei and Zhu, Lei and Xu, Xuemiao and Qin, Jing and Han, Guoqiang and Heng, Pheng-Ann},
	booktitle={Proceedings of the 27th international joint conference on artificial intelligence},
	volume={684690},
	year={2018},
	organization={AAAI Press Menlo Park, CA, USA}
}

@inproceedings{liu2019simple,
	title={A simple pooling-based design for real-time salient object detection},
	author={Liu, Jiang-Jiang and Hou, Qibin and Cheng, Ming-Ming and Feng, Jiashi and Jiang, Jianmin},
	booktitle={Proceedings of the IEEE/CVF conference on computer vision and pattern recognition},
	pages={3917--3926},
	year={2019}
}

@inproceedings{gao2020highly,
	title={Highly efficient salient object detection with 100k parameters},
	author={Gao, Shang-Hua and Tan, Yong-Qiang and Cheng, Ming-Ming and Lu, Chengze and Chen, Yunpeng and Yan, Shuicheng},
	booktitle={European conference on computer vision},
	pages={702--721},
	year={2020},
	organization={Springer}
}

@article{li2020stacked,
	title={Stacked U-shape network with channel-wise attention for salient object detection},
	author={Li, Junxia and Pan, Zefeng and Liu, Qingshan and Wang, Ziyang},
	journal={IEEE Transactions on Multimedia},
	volume={23},
	pages={1397--1409},
	year={2020},
	publisher={IEEE}
}

@inproceedings{xu2021locate,
	title={Locate globally, segment locally: A progressive architecture with knowledge review network for salient object detection},
	author={Xu, Binwei and Liang, Haoran and Liang, Ronghua and Chen, Peng},
	booktitle={Proceedings of the AAAI conference on artificial intelligence},
	volume={35},
	number={4},
	pages={3004--3012},
	year={2021}
}

@inproceedings{liu2021visual,
	title={Visual saliency transformer},
	author={Liu, Nian and Zhang, Ni and Wan, Kaiyuan and Shao, Ling and Han, Junwei},
	booktitle={Proceedings of the IEEE/CVF international conference on computer vision},
	pages={4722--4732},
	year={2021}
}

@article{liu2022disentangled,
	title={Disentangled capsule routing for fast part-object relational saliency},
	author={Liu, Yi and Zhang, Dingwen and Liu, Nian and Xu, Shoukun and Han, Jungong},
	journal={IEEE Transactions on Image Processing},
	volume={31},
	pages={6719--6732},
	year={2022},
	publisher={IEEE}
}

@article{fang2022densely,
	title={Densely nested top-down flows for salient object detection},
	author={Fang, Chaowei and Tian, Haibin and Zhang, Dingwen and Zhang, Qiang and Han, Jungong and Han, Junwei},
	journal={Science China Information Sciences},
	volume={65},
	number={8},
	pages={182103},
	year={2022},
	publisher={Springer}
}

@article{zhuge2022salient,
	title={Salient object detection via integrity learning},
	author={Zhuge, Mingchen and Fan, Deng-Ping and Liu, Nian and Zhang, Dingwen and Xu, Dong and Shao, Ling},
	journal={IEEE Transactions on Pattern Analysis and Machine Intelligence},
	volume={45},
	number={3},
	pages={3738--3752},
	year={2022},
	publisher={IEEE}
}

@article{li2019nested,
	title={Nested network with two-stream pyramid for salient object detection in optical remote sensing images},
	author={Li, Chongyi and Cong, Runmin and Hou, Junhui and Zhang, Sanyi and Qian, Yue and Kwong, Sam},
	journal={IEEE Transactions on Geoscience and Remote Sensing},
	volume={57},
	number={11},
	pages={9156--9166},
	year={2019},
	publisher={IEEE}
}

@article{tu2021orsi,
	title={ORSI salient object detection via multiscale joint region and boundary model},
	author={Tu, Zhengzheng and Wang, Chao and Li, Chenglong and Fan, Minghao and Zhao, Haifeng and Luo, Bin},
	journal={IEEE Transactions on Geoscience and Remote Sensing},
	volume={60},
	pages={1--13},
	year={2021},
	publisher={IEEE}
}

@article{wang2022multiscale,
	title={Multiscale feature enhancement network for salient object detection in optical remote sensing images},
	author={Wang, Zhen and Guo, Jianxin and Zhang, Chuanlei and Wang, Buhong},
	journal={IEEE Transactions on Geoscience and Remote Sensing},
	volume={60},
	pages={1--19},
	year={2022},
	publisher={IEEE}
}

@article{li2022adjacent,
	title={Adjacent context coordination network for salient object detection in optical remote sensing images},
	author={Li, Gongyang and Liu, Zhi and Zeng, Dan and Lin, Weisi and Ling, Haibin},
	journal={IEEE Transactions on Cybernetics},
	volume={53},
	number={1},
	pages={526--538},
	year={2022},
	publisher={IEEE}
}

@article{li2021multi,
	title={Multi-content complementation network for salient object detection in optical remote sensing images},
	author={Li, Gongyang and Liu, Zhi and Lin, Weisi and Ling, Haibin},
	journal={IEEE Transactions on Geoscience and Remote Sensing},
	volume={60},
	pages={1--13},
	year={2021},
	publisher={IEEE}
}

@article{wang2022hybrid,
	title={Hybrid feature aligned network for salient object detection in optical remote sensing imagery},
	author={Wang, Qi and Liu, Yanfeng and Xiong, Zhitong and Yuan, Yuan},
	journal={IEEE transactions on geoscience and remote sensing},
	volume={60},
	pages={1--15},
	year={2022},
	publisher={IEEE}
}

@article{liu2024heterogeneous,
	title={Heterogeneous Feature Collaboration Network for Salient Object Detection in Optical Remote Sensing Images},
	author={Liu, Yutong and Xu, Mingzhu and Xiao, Tianxiang and Tang, Haoyu and Hu, Yupeng and Nie, Liqiang},
	journal={IEEE Transactions on Geoscience and Remote Sensing},
	year={2024},
	publisher={IEEE}
}

@article{ren2022ship,
	title={Ship detection in high-resolution optical remote sensing images aided by saliency information},
	author={Ren, Zhida and Tang, Yongqiang and He, Zewen and Tian, Lei and Yang, Yang and Zhang, Wensheng},
	journal={IEEE Transactions on Geoscience and Remote Sensing},
	volume={60},
	pages={1--16},
	year={2022},
	publisher={IEEE}
}

@article{yao2024iterative,
	title={Iterative saliency aggregation and assignment network for efficient salient object detection in optical remote sensing images},
	author={Yao, Zhaojian and Gao, Wei},
	journal={IEEE Transactions on Geoscience and Remote Sensing},
	year={2024},
	publisher={IEEE}
}

@inproceedings{ma2003contrast,
	title={Contrast-based image attention analysis by using fuzzy growing},
	author={Ma, Yu-Fei and Zhang, Hong-Jiang},
	booktitle={Proceedings of the eleventh ACM international conference on Multimedia},
	pages={374--381},
	year={2003}
}

@article{han2006unsupervised,
	title={Unsupervised extraction of visual attention objects in color images},
	author={Han, Junwei and Ngan, King Ngi and Li, Mingjing and Zhang, Hong-Jiang},
	journal={IEEE transactions on circuits and systems for video technology},
	volume={16},
	number={1},
	pages={141--145},
	year={2006},
	publisher={IEEE}
}

@article{li2020complementarity,
	title={Complementarity-aware attention network for salient object detection},
	author={Li, Junxia and Pan, Zefeng and Liu, Qingshan and Cui, Ying and Sun, Yubao},
	journal={IEEE transactions on cybernetics},
	volume={52},
	number={2},
	pages={873--886},
	year={2020},
	publisher={IEEE}
}

@article{fang2023udnet,
	title={UDNet: Uncertainty-aware deep network for salient object detection},
	author={Fang, Yuming and Zhang, Haiyan and Yan, Jiebin and Jiang, Wenhui and Liu, Yang},
	journal={Pattern recognition},
	volume={134},
	pages={109099},
	year={2023},
	publisher={Elsevier}
}

@article{lu2024low,
	title={Low-light salient object detection by learning to highlight the foreground objects},
	author={Lu, Xiao and Yuan, Yulin and Liu, Xing and Wang, Lucai and Zhou, Xuanyu and Yang, Yimin},
	journal={IEEE Transactions on Circuits and Systems for Video Technology},
	year={2024},
	publisher={IEEE}
}

@article{liu2023distilling,
	title={Distilling knowledge from super-resolution for efficient remote sensing salient object detection},
	author={Liu, Yanfeng and Xiong, Zhitong and Yuan, Yuan and Wang, Qi},
	journal={IEEE Transactions on Geoscience and Remote Sensing},
	volume={61},
	pages={1--16},
	year={2023},
	publisher={IEEE}
}

@article{yan2024asnet,
	title={ASNet: Adaptive semantic network based on transformer--CNN for salient object detection in optical remote sensing images},
	author={Yan, Ruixiang and Yan, Longquan and Geng, Guohua and Cao, Yufei and Zhou, Pengbo and Meng, Yongle},
	journal={IEEE Transactions on Geoscience and Remote Sensing},
	volume={62},
	pages={1--16},
	year={2024},
	publisher={IEEE}
}

@article{zhao2024recurrent,
	title={Recurrent adaptive graph reasoning network with region and boundary interaction for salient object detection in optical remote sensing images},
	author={Zhao, Jie and Jia, Yun and Ma, Lin and Yu, Lidan},
	journal={IEEE Transactions on Geoscience and Remote Sensing},
	year={2024},
	publisher={IEEE}
}

@article{chen2025replay,
	title={Replay Without Saving: Prototype Derivation and Distribution Rebalance for Class-Incremental Semantic Segmentation},
	author={Chen, Jinpeng and Cong, Runmin and Luo, Yuxuan and Ip, Horace Ho Shing and Kwong, Sam},
	journal={IEEE transactions on pattern analysis and machine intelligence},
	year={2025},
	publisher={IEEE}
}

@article{dosovitskiy2020image,
	title={An image is worth 16x16 words: Transformers for image recognition at scale},
	author={Dosovitskiy, Alexey and Beyer, Lucas and Kolesnikov, Alexander and Weissenborn, Dirk and Zhai, Xiaohua and Unterthiner, Thomas and Dehghani, Mostafa and Minderer, Matthias and Heigold, Georg and Gelly, Sylvain and others},
	journal={arXiv preprint arXiv:2010.11929},
	year={2020}
}

@inproceedings{wang2021pyramid,
	title={Pyramid vision transformer: A versatile backbone for dense prediction without convolutions},
	author={Wang, Wenhai and Xie, Enze and Li, Xiang and Fan, Deng-Ping and Song, Kaitao and Liang, Ding and Lu, Tong and Luo, Ping and Shao, Ling},
	booktitle={Proceedings of the IEEE/CVF international conference on computer vision},
	pages={568--578},
	year={2021}
}

@article{zhao2024adaptive,
	title={Adaptive dual-stream sparse transformer network for salient object detection in optical remote sensing images},
	author={Zhao, Jie and Jia, Yun and Ma, Lin and Yu, Lidan},
	journal={IEEE Journal of Selected Topics in Applied Earth Observations and Remote Sensing},
	volume={17},
	pages={5173--5192},
	year={2024},
	publisher={IEEE}
}

@article{gao2023adaptive,
	title={Adaptive spatial tokenization transformer for salient object detection in optical remote sensing images},
	author={Gao, Lina and Liu, Bing and Fu, Ping and Xu, Mingzhu},
	journal={IEEE Transactions on Geoscience and Remote Sensing},
	volume={61},
	pages={1--15},
	year={2023},
	publisher={IEEE}
}

@article{cong2025generalized,
	title={Generalized few-shot segmentation for remote sensing image based on class relation mining},
	author={Cong, Runmin and Sun, Haoyan and Luo, Yuxuan and Fang, Hao},
	journal={Acta Aeronautica et Astronautica Sinica},
	volume={46},
	number={23},
	year={2025},
	publisher={航空学报杂志社}
}

@article{cong2025breaking,
	title={Breaking Barriers, Localizing Saliency: A Large-scale Benchmark and Baseline for Condition-Constrained Salient Object Detection},
	author={Cong, Runmin and Chen, Zhiyang and Fang, Hao and Kwong, Sam and Zhang, Wei},
	journal={IEEE Transactions on Pattern Analysis and Machine Intelligence},
	year={2025},
	publisher={IEEE}
}

@article{qin2025sight,
	title={From Sight to Insight: Unleashing Eye-Tracking in Weakly Supervised Video Salient Object Detection},
	author={Qin, Qi and Cong, Runmin and Zhan, Gen and Liao, Yiting and Kwong, Sam},
	journal={IEEE Transactions on Multimedia},
	year={2025},
	publisher={IEEE}
}

@article{cong2025divide,
	title={Divide-and-Conquer Decoupled Network for Cross-Domain Few-Shot Segmentation},
	author={Cong, Runmin and Wang, Anpeng and Wan, Bin and Zhang, Cong and Zhou, Xiaofei and Zhang, Wei},
	journal={arXiv preprint arXiv:2511.07798},
	year={2025}
}

@article{chen2025empowering,
	title={Empowering DINO Representations for Underwater Instance Segmentation via Aligner and Prompter},
	author={Chen, Zhiyang and Zhang, Chen and Fang, Hao and Cong, Runmin},
	journal={arXiv preprint arXiv:2511.08334},
	year={2025}
}

@inproceedings{xiong2025mm,
	title={MM-Prompt: Multi-modality and Multi-granularity Prompts for Few-Shot Segmentation},
	author={Xiong, Hang and Cong, Runmin and Chen, Jinpeng and Zhang, Chen and Li, Feng and Bai, Huihui and Kwong, Sam},
	booktitle={Proceedings of the 33rd ACM International Conference on Multimedia},
	pages={3067--3075},
	year={2025}
}

@inproceedings{cong2025uis,
	title={UIS-Mamba: exploring mamba for underwater instance segmentation via dynamic tree scan and hidden state weaken},
	author={Cong, Runmin and Yu, Zongji and Fang, Hao and Sun, Haoyan and Kwong, Sam},
	booktitle={Proceedings of the 33rd ACM International Conference on Multimedia},
	pages={343--352},
	year={2025}
}

@article{cong2025reference,
	title={Reference-based iterative interaction with p 2-matching for stereo image super-resolution},
	author={Cong, Runmin and Liao, Rongxin and Li, Feng and Sheng, Ronghui and Bai, Huihui and Wan, Renjie and Kwong, Sam and Zhang, Wei},
	journal={IEEE Transactions on Image Processing},
	year={2025},
	publisher={IEEE}
}

@article{cong2025trnet,
	title={Trnet: Two-tier recursion network for co-salient object detection},
	author={Cong, Runmin and Yang, Ning and Liu, Hongyu and Zhang, Dingwen and Huang, Qingming and Kwong, Sam and Zhang, Wei},
	journal={IEEE Transactions on Circuits and Systems for Video Technology},
	volume={35},
	number={6},
	pages={5844--5857},
	year={2025},
	publisher={IEEE}
}

@inproceedings{fang2025decoupled,
	title={Decoupled motion expression video segmentation},
	author={Fang, Hao and Cong, Runmin and Lu, Xiankai and Zhou, Xiaofei and Kwong, Sam and Zhang, Wei},
	booktitle={Proceedings of the Computer Vision and Pattern Recognition Conference},
	pages={13821--13831},
	year={2025}
}

@article{cong2024query,
	title={Query-guided prototype evolution network for few-shot segmentation},
	author={Cong, Runmin and Xiong, Hang and Chen, Jinpeng and Zhang, Wei and Huang, Qingming and Zhao, Yao},
	journal={IEEE Transactions on Multimedia},
	volume={26},
	pages={6501--6512},
	year={2024},
	publisher={IEEE}
}

@article{li2016single,
	title={Single underwater image enhancement based on color cast removal and visibility restoration},
	author={Li, Chongyi and Guo, Jichang and Wang, Bo and Cong, Runmin and Zhang, Yan and Wang, Jian},
	journal={Journal of Electronic Imaging},
	volume={25},
	number={3},
	pages={033012--033012},
	year={2016},
	publisher={Society of Photo-Optical Instrumentation Engineers}
}

@article{cong2022bcs,
	title={BCS-Net: Boundary, context, and semantic for automatic COVID-19 lung infection segmentation from CT images},
	author={Cong, Runmin and Yang, Haowei and Jiang, Qiuping and Gao, Wei and Li, Haisheng and Wang, Cong and Zhao, Yao and Kwong, Sam},
	journal={IEEE Transactions on Instrumentation and Measurement},
	volume={71},
	pages={1--11},
	year={2022},
	publisher={IEEE}
}

@article{lian2024diving,
	title={Diving into underwater: Segment anything model guided underwater salient instance segmentation and a large-scale dataset},
	author={Lian, Shijie and Zhang, Ziyi and Li, Hua and Li, Wenjie and Yang, Laurence Tianruo and Kwong, Sam and Cong, Runmin},
	journal={arXiv preprint arXiv:2406.06039},
	year={2024}
}

@inproceedings{jing2021occlusion,
	title={Occlusion-aware bi-directional guided network for light field salient object detection},
	author={Jing, Dong and Zhang, Shuo and Cong, Runmin and Lin, Youfang},
	booktitle={Proceedings of the 29th ACM international conference on multimedia},
	pages={1692--1701},
	year={2021}
}
\vspace{-4\baselineskip}
\end{document}